\documentclass[lettersize,journal]{IEEEtran}
\usepackage{amsmath,amsfonts}
\usepackage{algorithmic}
\usepackage{algorithm}
\usepackage{array}
\usepackage[caption=false,font=normalsize,labelfont=sf,textfont=sf]{subfig}
\usepackage{textcomp}
\usepackage{stfloats}
\usepackage{url}
\usepackage{verbatim}
\usepackage{graphicx}
\usepackage{cite}
\usepackage{enumitem}
\newcommand{\modelname}{GPOcc}
\newcommand{\newmodelname}{GPOcc++}

\usepackage[table]{xcolor}
\definecolor{impblue}{RGB}{64,150,230}
\newcommand{\gain}[1]{\textcolor{impblue}{#1}}

\usepackage{multirow}

\usepackage{colortbl}

\usepackage{booktabs}

\usepackage{bm}

\usepackage{graphicx}

\usepackage{caption}

\usepackage{array}

\usepackage{amssymb}

\usepackage{amsmath}

\usepackage{wrapfig}

\usepackage{placeins}

\usepackage{booktabs,makecell,multirow,pifont}

\usepackage[hidelinks]{hyperref}
\usepackage{cleveref}

\begin{document}

\title{GPOcc++: Unified Sparse Gaussian Occupancy Prediction with Visual Geometry Priors}

\author{Changqing Zhou,
Yueru Luo,
Yulan Guo,
Bing Wang,
Jie Qin,
and Changhao Chen%
\thanks{Changqing Zhou and Changhao Chen are with The Hong Kong University
of Science and Technology (Guangzhou), Guangzhou 511453, China
(e-mail: czhou149@connect.hkust-gz.edu.cn;
changhaochen@hkust-gz.edu.cn). Changhao Chen is also with the Division of
Emerging Interdisciplinary Areas, The Hong Kong University of Science and
Technology, Hong Kong SAR 999077, China.}%
\thanks{Yueru Luo is with The Chinese University of Hong Kong, Shenzhen,
China.}%
\thanks{Yulan Guo is with the School of Electronics and Communication
Engineering, Sun Yat-sen University, Shenzhen 518107, China.}%
\thanks{Bing Wang is with the Faculty of Engineering, The Hong Kong
Polytechnic University, Hong Kong SAR, China.}%
\thanks{Jie Qin is with the College of Artificial Intelligence, Nanjing
University of Aeronautics and Astronautics, Nanjing 211106, China.}%
\thanks{Corresponding author: Changhao Chen.}%
}


\maketitle

\begin{figure*}[t]
    \centering
    \scriptsize
    \newcolumntype{P}[1]{>{\centering\arraybackslash}p{#1}}
    \setlength{\tabcolsep}{0.01\linewidth} %
    \renewcommand{\arraystretch}{1.5}      %
    \begin{tabular}{P{0.31\textwidth} P{0.31\textwidth} P{0.31\textwidth}}
        \includegraphics[width=0.95\linewidth]{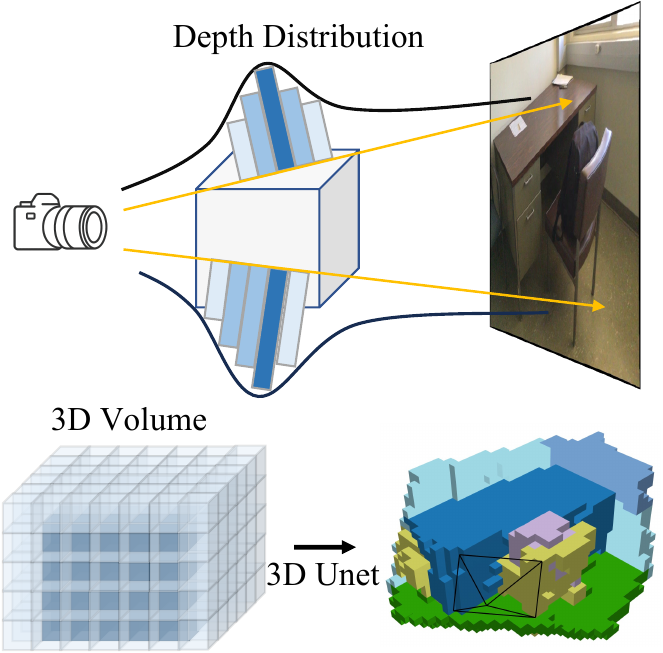} &  
        \includegraphics[width=0.95\linewidth]{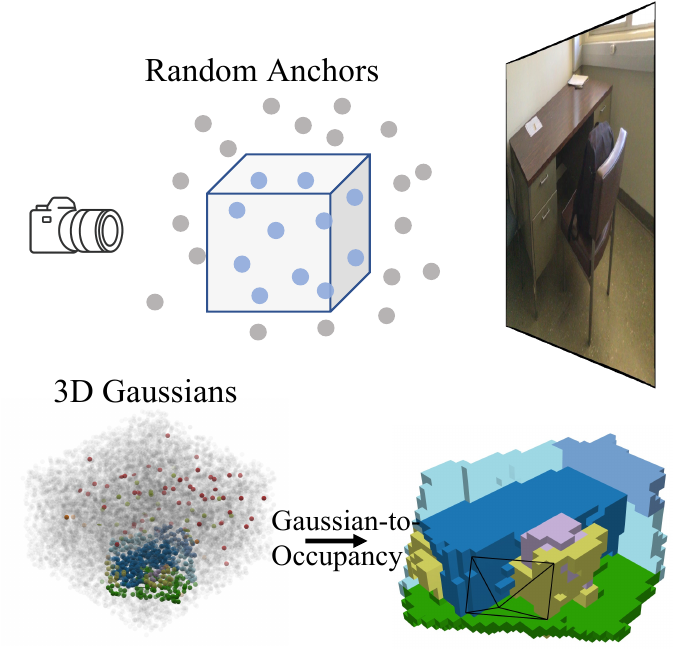} &  
        \includegraphics[width=0.95\linewidth]{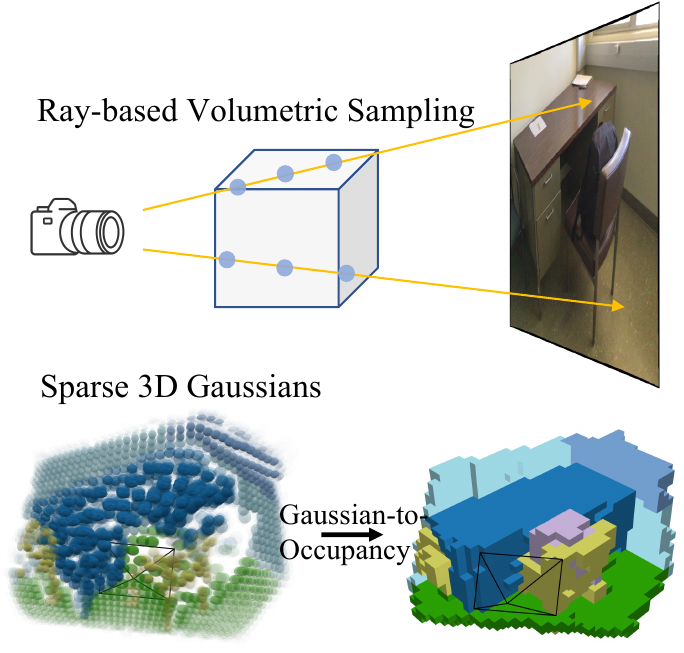} \\
        (a) ISO~\cite{ISO} & (b) EmbodiedOcc~\cite{embodiedocc} & (c) Ours
    \end{tabular}
\caption{\textbf{Comparison of monocular occupancy prediction pipelines.} 
ISO~\cite{ISO} formulates depth estimation as a multi-class classification problem, using the predicted depth distributions to lift 2D image features into dense 3D volumes, which are then processed by a 3D U-Net for occupancy prediction. 
EmbodiedOcc~\cite{embodiedocc}, by contrast, initializes random 3D anchors and applies cross-attention to aggregate image features, predicting Gaussian primitives that are splatted into voxels. Many of these Gaussians fall in empty regions, shown as gray primitives. 
In contrast, \modelname~employs ray-based volumetric sampling to generate sparse Gaussians concentrated on or within objects, producing a compact and efficient representation for occupancy.}
    \label{fig:teaser}
\end{figure*}

\begin{abstract}
Accurate 3D scene understanding is fundamental to embodied intelligence and autonomous driving, where 3D occupancy provides a unified representation of objects, structures, and free space.
However, recovering such a complete volumetric representation from visual observations remains challenging, particularly in occluded and unobserved regions. Visual geometry priors offer strong and generalizable geometric cues for addressing this challenge, but their outputs are inherently surface-centric, whereas occupancy prediction requires reasoning about volumetric interiors and free space.
To bridge this gap, we introduce GPOcc, which transforms visual geometry priors into occupancy-aware sparse Gaussian representations for efficient and expressive volumetric scene modeling. Building on GPOcc, GPOcc++ models multi-view observations and temporal sequences within a unified framework, allowing spatial and temporal evidence to be handled through the same representation.
We further extend GPOcc++ from indoor scenes to outdoor occupancy prediction. Extensive experiments on both indoor and outdoor benchmarks demonstrate consistently strong performance across both multi-view and temporal settings, together with favorable efficiency and generalization.
Code will be released at \url{https://github.com/JuIvyy/GPOcc}.
\end{abstract}


\definecolor{ceiling}{RGB}{214,  38, 40}   %
\definecolor{floor}{RGB}{43, 160, 4}     %
\definecolor{wall}{RGB}{158, 216, 229}  %
\definecolor{window}{RGB}{114, 158, 206}  %
\definecolor{chair}{RGB}{204, 204, 91}   %
\definecolor{bed}{RGB}{255, 186, 119}  %
\definecolor{sofa}{RGB}{147, 102, 188}  %
\definecolor{table}{RGB}{30, 119, 181}   %
\definecolor{tvs}{RGB}{160, 188, 33}   %
\definecolor{furniture}{RGB}{255, 127, 12}  %
\definecolor{objects}{RGB}{196, 175, 214} %

\definecolor{nbarrier}{RGB}{255, 120, 50}
\definecolor{nbicycle}{RGB}{255, 192, 203}
\definecolor{nbus}{RGB}{255, 255, 0}
\definecolor{ncar}{RGB}{0, 150, 245}
\definecolor{nconstruct}{RGB}{0, 255, 255}
\definecolor{nmotor}{RGB}{200, 180, 0}
\definecolor{npedestrian}{RGB}{255, 0, 0}
\definecolor{ntraffic}{RGB}{255, 240, 150}
\definecolor{ntrailer}{RGB}{135, 60, 0}
\definecolor{ntruck}{RGB}{160, 32, 240}
\definecolor{ndriveable}{RGB}{255, 0, 255}
\definecolor{nother}{RGB}{139, 137, 137}
\definecolor{nsidewalk}{RGB}{75, 0, 75}
\definecolor{nterrain}{RGB}{150, 240, 80}
\definecolor{nmanmade}{RGB}{213, 213, 213}
\definecolor{nvegetation}{RGB}{0, 175, 0}

\section{Introduction}

\IEEEPARstart{E}{mbodied} AI agents and autonomous systems increasingly require accurate and detailed 3D understanding of their surroundings~\cite{embodiedaisurvey2025}, which forms the foundation for reasoning, planning, and interaction in complex environments. Recent advances in vision-based perception have significantly improved 3D scene understanding by exploiting rich semantic and geometric cues~\cite{bevformer,unimode,indoordepth,imvoxelnet,occfor3ddet}. Among various scene representations, 3D occupancy~\cite{ISO,embodiedocc} has emerged as a powerful paradigm by jointly modeling foreground objects, background structures, and free space within a unified volumetric representation, making it a fundamental component for downstream tasks such as robotic navigation~\cite{volumetrivln}, manipulation~\cite{occvla}, and autonomous driving~\cite{occllama}.

Although vision-based occupancy estimation has been extensively investigated in autonomous driving~\cite{voxformer,surroundocc,Triformer,occupancypoints,opus,gaussianformer,gaussianformer2}, fine-grained occupancy prediction in indoor environments remains relatively underexplored. Compared with outdoor driving scenes, indoor environments exhibit highly cluttered spatial layouts, substantial appearance variability, and a broader diversity of object categories, posing additional challenges for accurate volumetric reasoning. Recent approaches such as ISO~\cite{ISO} and EmbodiedOcc~\cite{embodiedocc} have demonstrated promising results by leveraging depth priors~\cite{depthanything,depthanythingv2}. Specifically, ISO lifts image features into dense volumetric representations through estimated depth distributions and performs occupancy reasoning using a 3D U-Net. EmbodiedOcc, in contrast, initializes Gaussian primitives and iteratively refines them through cross-attention between projected 3D anchors and image features before splatting the refined Gaussians into occupancy space. Despite their effectiveness, these methods exploit only limited geometric cues~\cite{yang2025bevheight++} and often devote substantial representational capacity to large empty regions, leading to reduced efficiency and constrained generalization.

Meanwhile, recent progress in \emph{visual geometry priors} has introduced powerful geometric representations, ranging from monocular depth foundation models such as the DepthAnything family~\cite{depthanything,depthanythingv2} to multi-view Visual Geometry Models (VGMs)~\cite{vggt,dust3r,mast3r,fast3r,pi3,point3r}. These models provide rich geometric information, including depth maps, point maps, and camera parameters, enabling high-quality 3D reconstruction across diverse environments. However, such representations are inherently surface-centric: each image pixel typically corresponds to a single visible surface point, whereas occupancy prediction requires reasoning over occupied and free space throughout a 3D volume, including occluded and unobserved regions. This discrepancy raises a fundamental question: \emph{how can surface-centric visual geometry priors be transformed into volumetric scene representations suitable for occupancy reasoning?}

To address this challenge, we proposed \textbf{\modelname}, a framework that combines generalizable \textbf{G}eometry \textbf{P}riors (GPs) with sparse Gaussian representations for fine-grained 3D occupancy prediction~\cite{gpocc}. Sparse Gaussians provide continuous spatial support around geometry-prior points while maintaining a compact representation, avoiding the unnecessary computation introduced by densely modeling empty space. Their probabilistic formulation also enables differentiable aggregation and naturally accommodates uncertainty in the inferred volumetric support.
Our approach builds on four key components. \textbf{(1)} To overcome the surface-only nature of geometry priors, a \emph{ray-based volumetric sampling} strategy extends geometry-prior points along camera rays to generate volumetric anchors, as illustrated in~\Cref{fig:teaser}(c). Each anchor predicts a Gaussian primitive that models its local spatial neighborhood. \textbf{(2)} An \emph{opacity-based pruning} mechanism removes low-contribution Gaussians, substantially reducing redundancy while preserving informative volumetric support. \textbf{(3)} Occupancy is estimated from the remaining sparse Gaussian set through a probabilistic formulation following~\cite{gaussianformer2}, transforming continuous Gaussian support into discrete occupancy predictions. \textbf{(4)} For embodied applications with streaming observations, a training-free \emph{incremental update strategy} progressively fuses per-frame Gaussian representations into a global scene representation. Together, these components establish a principled transformation from surface-centric geometry priors to volumetric representations.

While \modelname~demonstrates the effectiveness of geometry priors for occupancy estimation, several limitations remain. First, it processes geometry priors largely in a frame-wise manner and therefore cannot explicitly exploit complementary temporal or multi-view observations to resolve ambiguities caused by occlusion, missing evidence, and limited fields of view. Second, its fixed ray-sampled anchors provide limited flexibility for modeling irregular shapes, thin structures, object boundaries, and local geometry errors. Third, \modelname~has mainly been validated in indoor embodied environments, leaving its generalization to large-scale outdoor autonomous-driving scenarios insufficiently explored.

Motivated by these observations, we further propose \textbf{\newmodelname}, a substantially enhanced framework that extends \modelname~in both capability and applicability. \newmodelname~introduces two key technical advances while broadening geometry-prior-based occupancy prediction from indoor embodied environments to large-scale autonomous-driving scenarios.
First, we propose \textit{Ray-Conditioned Multi-Image Fusion}, which performs geometry-aware fusion across observations acquired from different viewpoints and timestamps within a unified representation space. By explicitly modeling camera-ray geometry, the proposed module exploits complementary observations to reduce ambiguities in single-view geometry priors while naturally supporting single-frame, temporal, and multi-view inputs. 
Second, we introduce \textit{Offset-Guided Ray Anchoring}, which predicts explicit 3D offsets for ray-sampled anchors, allowing Gaussian primitives to adaptively align with complex scene structures while preserving the geometric inductive bias of camera rays. Together, these advances enable more accurate and flexible volumetric scene representations across diverse environments.
Extensive experiments on indoor embodied-scene and outdoor autonomous-driving benchmarks demonstrate that \newmodelname~consistently achieves state-of-the-art or highly competitive performance under single-frame, temporal, and multi-view settings while maintaining favorable computational efficiency. These results demonstrate the effectiveness and generality of the proposed geometry-prior-based surface-to-volume transformation across diverse environments.

Compared with GPOcc, the additional contributions of this work are summarized as follows:

\begin{enumerate}[leftmargin=5.8mm]
\item We propose \newmodelname, a geometry-prior-based framework for 3D occupancy prediction that extends surface-to-volume transformation from frame-wise processing to unified temporal and multi-view reasoning.

\item We introduce \textit{Ray-Conditioned Multi-Image Fusion}, which explicitly models cross-view and cross-temporal geometric relationships and exploits complementary observations to alleviate ambiguities in single-image geometry priors.

\item We introduce \textit{Offset-Guided Ray Anchoring}, which predicts explicit 3D residuals for ray-sampled anchors, enabling Gaussian centers to better align with complex scene structures.

\item We extend geometry-prior-based occupancy prediction from embodied environments to autonomous-driving scenarios and conduct extensive experiments demonstrating state-of-the-art or highly competitive performance with favorable efficiency.

\end{enumerate}

This article builds upon our previous conference paper~\cite{gpocc} published at CVPR 2026. Compared with the conference version, the present work introduces \textit{Ray-Conditioned Multi-Image Fusion} for unified multi-view and multi-frame reasoning, \textit{Offset-Guided Ray Anchoring} for more flexible geometric modeling, and comprehensive validation across both indoor and outdoor benchmarks, accompanied by substantially expanded experimental analyses.

\section{Related Work}

\subsection{Visual Geometry Foundation Models}

Earlier geometry learning methods have explored generalizable monocular depth~\cite{ranftl2020towards} estimation and ray-based~\cite{shi2023raymvsnet++} multi-view reconstruction. Despite their strong geometric reasoning capabilities, these methods primarily recover depth or visible surface geometry and do not directly construct volumetric representations required by occupancy prediction.
More recently, visual geometry foundation models have learned more unified geometric representations from large-scale data. DUSt3R~\cite{dust3r} and MASt3R~\cite{mast3r} predict coupled scene representations, including camera poses and geometry parameterized by pointmaps, from image pairs. However, they still require expensive post-processing or symmetric inference for unconstrained multi-view SfM. Subsequent works including Spann3R~\cite{spann3r}, CUT3R~\cite{cut3r}, and MUSt3R~\cite{must3r} reduce the reliance on classical optimization by introducing latent-state memory in transformers, enabling more end-to-end multi-view reconstruction. Fast3R~\cite{fast3r} further scales this paradigm to efficiently handle more than 1000 input images.
Building on this line, VGGT~\cite{vggt} jointly predicts pointmaps, depth, camera poses, and tracking features with minimal hand-crafted 3D inductive biases. StreamVGGT~\cite{streamvggt} reformulates VGGT with a causal transformer for efficient long-sequence processing. Several variants have also been developed on top of VGGT: $\pi^3$~\cite{pi3} removes the dependence on the first input frame as the reference coordinate system, while Dens3R~\cite{dens3r} enriches geometric predictions with surface normals.
These models provide increasingly strong priors for downstream 3D understanding. Nevertheless, their outputs remain largely surface-centric, typically describing one visible surface point per image pixel. In contrast, our work studies how such general visual geometry priors can be transformed into sparse volumetric representations that model both visible surfaces and latent occupancy support.

\subsection{Occupancy Prediction}

MonoScene~\cite{monoscene} extends semantic scene completion~\cite{monoscene,ndcscene,sscnet} to monocular 3D occupancy prediction. Earlier semantic scene completion methods have explored anisotropic 3D convolutions~\cite{li2021anisotropic}, local implicit functions~\cite{rist2021semantic}, and semantic reconstruction from posed RGB images~\cite{huang2024ssr}. These approaches demonstrate the importance of jointly recovering scene geometry and semantics, but typically rely on task-specific voxel, LiDAR, or implicit representations rather than general visual geometry priors.
Occupancy prediction has subsequently been extensively studied in autonomous-driving scenarios~\cite{ma2024vision,sparseocc,gaussianformer,gaussianformer2,voxformer}, with multi-camera BEV perception providing an important foundation for spatial reasoning~\cite{li2023delving}. Recent studies further extend occupancy beyond a single perception task: OccScene~\cite{li2025occscene} uses semantic occupancy to facilitate cross-task scene generation, while SPOT~\cite{yan2025spot} employs occupancy prediction as a scalable pre-training objective for transferable 3D representations. Large-scale datasets such as OmniHD-Scenes~\cite{zheng2026omnihd} further support multimodal autonomous-driving perception, and occupancy-like scene predictions have also been incorporated into integrated driving planning systems~\cite{liu2025hybrid}. In comparison, research on fine-grained indoor occupancy remains relatively limited~\cite{ISO,embodiedocc,embodiedocc++,legoocc,jiang2026freeocc}.
Existing occupancy methods adopt different scene representations and lifting strategies. Grid-based approaches lift 2D image features into dense volumetric spaces using depth distributions or ray-based transformations, followed by 3D convolutions or volumetric decoders~\cite{monoscene,ndcscene,ISO,fbocc,lss}. Transformer-based architectures have also been introduced for volumetric representation learning~\cite{occupancypoints}. Other methods inject depth or signed-distance cues into dense 3D grids~\cite{sscnet}, while tri-plane representations reduce the computational cost of full voxel volumes~\cite{Triformer}.
Beyond fixed grids, point- and Gaussian-based formulations initialize 3D anchors or primitives and refine them through iterative feature aggregation~\cite{gaussianformer,embodiedocc,opus}. Sparsity has also been exploited by pruning dense voxel features and subsequently processing the retained regions using sparse convolutions~\cite{sparseocc} or transformers~\cite{voxformer,octreeocc}. However, these methods generally construct their 3D representations from task-specific image features, predefined anchors, or dense intermediate volumes.

In contrast, our framework directly transforms surface-centric geometry priors into sparse Gaussian volumetric representations. It explicitly models cross-view and cross-temporal relationships via ray-conditioned feature fusion to exploit complementary observations and resolve ambiguities caused by occlusion and limited visibility. Furthermore, learnable 3D offsets enable ray-initialized Gaussian anchors to adapt to complex scene structures, providing a unified formulation for both indoor temporal and outdoor multi-camera inputs.

\section{Preliminaries}\label{sec:occ_gs_overview}
\noindent\textbf{Occupancy Prediction.}
Given one or more RGB images $\{\mathbf{I}_t\}_{t=1}^{T}$ with corresponding camera parameters, the goal is to predict a voxel-wise semantic occupancy map $\mathbf{O} \in \mathbb{R}^{X \times Y \times Z \times N_c}$, where $X$, $Y$, and $Z$ denote the spatial resolution of the scene and $N_c$ is the number of semantic classes. The case $T=1$ corresponds to monocular occupancy prediction, while $T>1$ represents temporal or multi-view inputs.

\noindent\textbf{Gaussian Representation.} A 3D scene can be represented by a set of semantic Gaussian primitives $\mathbf{G}=\{\mathcal{G}_i\}_{i=1}^{P}$. Each primitive $\mathcal{G}_i$ models a local spatial region centered at $\mu_i$ and is parameterized by scale $s_i$, rotation $r_i$, opacity $a_i$, and semantic feature $c_i$. This representation provides continuous spatial support while compactly encoding geometry and semantics.

\noindent\textbf{Gaussian-to-Occupancy.} Occupancy is obtained by aggregating the contributions of nearby Gaussian primitives for each voxel $p$~\cite{embodiedocc,gaussianformer,gaussianformer2}:
\begin{equation}\label{eq:gs_occ}
    \hat{o}(p;\mathbf{G})
    =
    \sum_{i\in\mathcal{N}(p)}
    g_i(p;\mu_i,s_i,r_i,a_i,c_i),
\end{equation}
where $\mathcal{N}(p)$ denotes the set of primitives influencing voxel $p$.

\section{\modelname}
\label{sec:gpocc}

\subsection{Overview}\label{sec:gpocc_overview}

\begin{figure*}[t]
  \centering
  \includegraphics[width=0.98\textwidth]{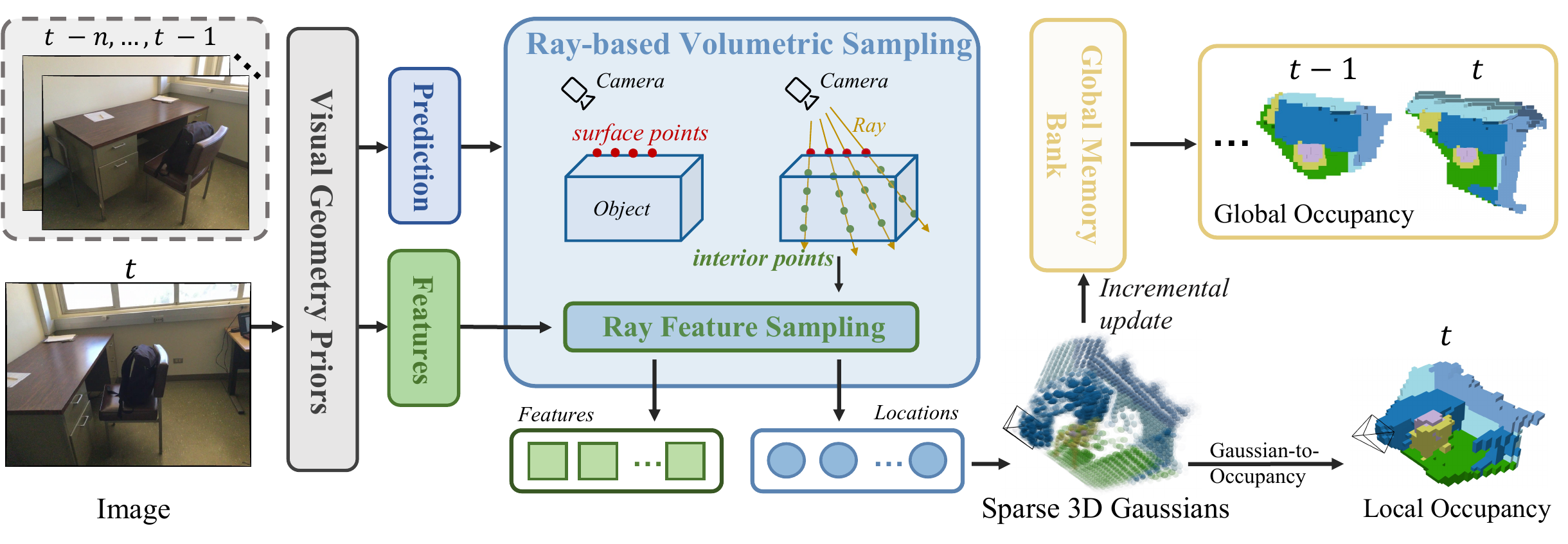}
  \vspace{-1mm}
  \caption{\textbf{Overview of \modelname.}
  Given an RGB image, a visual geometry prior predicts surface geometry and extracts 3D-aware features. Ray-based volumetric sampling extends the predicted surface points into interior volumetric anchors, which are represented by sparse semantic Gaussian primitives. The resulting Gaussians are probabilistically mapped to local occupancy. For streaming observations, per-frame Gaussian predictions are incrementally integrated into a global memory bank to reconstruct coherent scene-level occupancy.}
  \label{fig:framwork}
\end{figure*}

As illustrated in~\Cref{fig:framwork}, \modelname~uses a visual geometry prior to extract 3D-aware image features and surface-level geometric predictions. Ray-based volumetric sampling extends these visible surface points into volumetric anchors, from which semantic Gaussian primitives are predicted. Low-contribution primitives are removed through opacity pruning, and the remaining sparse Gaussians are probabilistically mapped to occupancy. For streaming observations, per-frame Gaussian predictions are further integrated into a global memory through a training-free incremental update strategy.

\subsection{Ray-based Volumetric Sampling}\label{sec:ray}

Visual geometry models~\cite{depthanythingv2,vggt,fast3r,point3r} provide strong priors by predicting depth or point maps. However, these predictions primarily describe visible surfaces, whereas occupancy reasoning also requires volumetric interiors and object thickness. We therefore extend surface predictions along their corresponding camera rays to construct volumetric anchors.

Given an RGB image $\mathbf{I}\in\mathbb{R}^{H\times W\times3}$, a geometry-prior model extracts an image feature map $\mathbf{F}\in\mathbb{R}^{H\times W\times C}$. To reduce computation, we downsample the intermediate features and regress a lower-resolution depth map:
\begin{equation}
    \mathbf{F}=\operatorname{GP}(\mathbf{I}),
    \mathbf{F}^{1/4}=\operatorname{DownSample}(\mathbf{F}),
    \mathbf{d}=\operatorname{MLP}_{\mathrm{depth}}(\mathbf{F}^{1/4})
\end{equation}
where $\mathbf{F}^{1/4} \in\mathbb{R}^{\frac{H}{4}\times\frac{W}{4}\times C}$ and $\mathbf{d}\in\mathbb{R}^{\frac{H}{4}\times\frac{W}{4}}$.

For a pixel $(u,v)$, its normalized camera-ray direction is computed from
the camera intrinsics:
\begin{equation}
    x=\frac{u-c_x}{f_x}, \qquad
    y=\frac{v-c_y}{f_y}, \qquad
    \mathbf{r}_{(u,v)}
    =
    \frac{[x,y,1]^\top}{\sqrt{x^2+y^2+1}},
\end{equation}
where $(c_x,c_y)$ and $(f_x,f_y)$ denote the principal point and focal lengths, respectively. The surface point predicted at $(u,v)$ is
\begin{equation}
    \mathbf{x}^{\mathrm{surf}}_{(u,v)}
    =
    \mathbf{d}_{(u,v)}\mathbf{r}_{(u,v)}.
\end{equation}
To approximate volumetric support beyond the visible surface, we sample
$K$ anchors along the same camera ray:
\begin{equation}
    \mathbf{x}_{(u,v,k)}
    =
    \bigl(\mathbf{d}_{(u,v)}+\delta_{(u,v,k)}\bigr)
    \mathbf{r}_{(u,v)}, \qquad k=1,\ldots,K,
\end{equation}
where the ray offsets are defined as
\begin{equation}\label{eq:learnable_scale}
    \{\delta_{(u,v,k)}\}_{k=1}^{K}
    =
    \operatorname{linspace}(0,1,K)
    \cdot
    \operatorname{scale}
    \bigl(\mathbf{F}^{1/4}_{(u,v)}\bigr).
\end{equation}
Here, $\operatorname{scale}(\cdot)$ is a lightweight predictor that adapts the ray extent to the local image and geometry features.

To distinguish the $K$ anchors associated with the same image location, we introduce a learnable embedding matrix $\mathbf{E}\in\mathbb{R}^{K\times C}$. The anchor-wise features are
constructed through broadcast addition:
\begin{equation}
    \hat{\mathbf{F}}^{1/4}
    =
    \mathbf{F}^{1/4}\oplus\mathbf{E},
    \qquad
    \hat{\mathbf{F}}^{1/4}
    \in
    \mathbb{R}^{\frac{H}{4}\times\frac{W}{4}\times K\times C},
\end{equation}
where $\oplus$ denotes broadcast addition. Gaussian attributes are then predicted for each anchor:
\begin{equation}
    \{s_i,r_i,a_i,c_i\}
    =
    \operatorname{MLP}_{\mathrm{gs}}
    \bigl(\hat{\mathbf{F}}^{1/4}\bigr),
\end{equation}
where $i=1,\ldots,\frac{H}{4}\times\frac{W}{4}\times K$, and the Gaussian center is initialized as the corresponding ray-sampled anchor $\mathbf{x}_{(u,v,k)}$.

By extending surface geometry along camera rays, \modelname~constructs volumetric support without introducing dense 3D anchors~\cite{embodiedocc} or lifting all image features into a dense voxel volume~\cite{ISO}.

\subsection{From Sparse Gaussians to Occupancy}\label{sec:gs2occ}

\begin{figure}[t]
  \centering
  \includegraphics[width=0.43\textwidth]{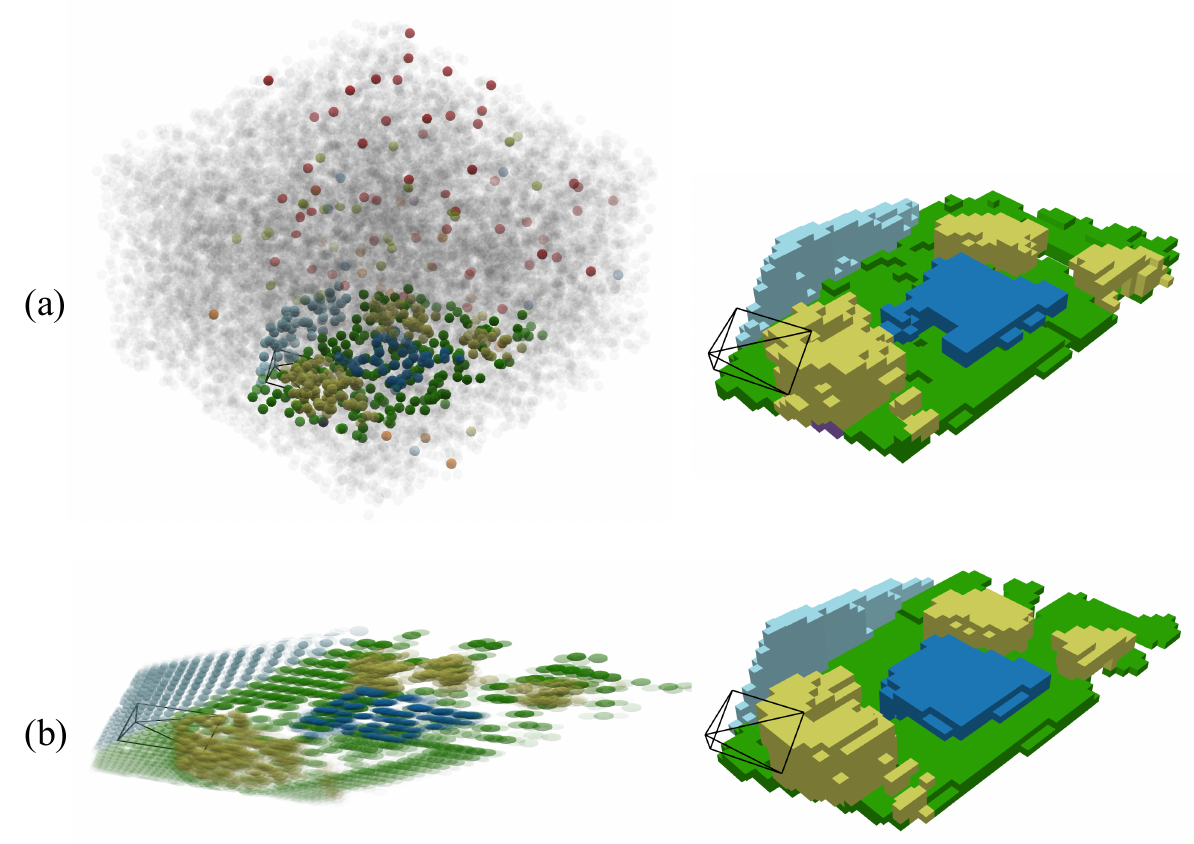}
  \vspace{-1mm}
  \caption{\textbf{Comparison of Gaussian representations.}
  (a) EmbodiedOcc places many Gaussian primitives in empty regions, shown
  in gray. (b) \modelname~concentrates sparse Gaussians on and within
  occupied structures, resulting in a more compact representation.}
  \label{fig:gs_compare}
\end{figure}

Prior Gaussian occupancy methods~\cite{embodiedocc} initialize a dense set of predefined 3D anchors and classify them as occupied or empty. Since most of the scene volume is unoccupied, many primitives are consequently assigned to empty regions, as shown in~\Cref{fig:gs_compare}(a). In contrast, ray-based volumetric sampling places Gaussian primitives near the predicted surface and interior regions, yielding a sparse distribution shown in~\Cref{fig:gs_compare}(b).
We adopt the probabilistic Gaussian superposition formulation of GaussianFormer-2~\cite{gaussianformer2}. For a query location $p$, the spatial contribution of Gaussian $\mathcal{G}_i$ is defined by
\begin{equation}
    o(p;\mathcal{G}_i)
    =
    \exp\left(
    -\frac{1}{2}
    (p-\mu_i)^\top
    \Sigma_i^{-1}
    (p-\mu_i)
    \right),
\end{equation}
where $\Sigma_i$ is constructed from the predicted scale $s_i$ and rotation $r_i$. Contributions from neighboring primitives are combined according to~\Cref{eq:gs_occ} to obtain voxel-wise semantic occupancy.
The Gaussian kernels define a continuous volumetric field whose spatial support is controlled by the predicted covariance. Under voxel-wise occupancy supervision, the Gaussian scales are optimized to match the corresponding object extent while leaving unsupported regions empty.
To further reduce redundancy, we discard primitives whose opacity satisfies $a_i<\tau$. We use $\tau=0.01$ by default, retaining only informative Gaussians for occupancy aggregation.

\subsection{Incremental Sparse Gaussian Update}\label{sec:online}

For embodied agents receiving streaming observations, \modelname~maintains a global Gaussian memory that progressively integrates per-frame predictions. Unlike methods that repeatedly update a dense predefined anchor set~\cite{embodiedocc}, our strategy directly accumulates sparse and adaptively generated primitives without additional training.

Let $\mathcal{M}$ denote the global Gaussian memory and $\mathbf{G}^{t}=\{\mathcal{G}^{t}_j\}_{j=1}^{P_t}$ the Gaussian set predicted from frame $\mathbf{I}_t$. Using the known camera pose, all Gaussians in $\mathbf{G}^{t}$ are first transformed from the camera coordinate system into the global coordinate system.

For each memory Gaussian $\mathcal{G}_i\in\mathcal{M}$, we identify its
newly observed neighbors
$\mathcal{N}(\mathcal{G}_i)\subseteq\mathbf{G}^{t}$ within a spatial radius
$\epsilon$. If neighboring primitives exist, their attributes are fused
through confidence-weighted averaging:
\begin{equation}
    \theta_i
    \leftarrow
    \frac{
    \gamma p_i\theta_i
    +(1-\gamma)
    \sum_{\mathcal{G}^{t}_j\in\mathcal{N}(\mathcal{G}_i)}
    p_j\theta_j
    }{
    \gamma p_i
    +(1-\gamma)
    \sum_{\mathcal{G}^{t}_j\in\mathcal{N}(\mathcal{G}_i)}
    p_j
    },
\end{equation}
where $\theta\in\{\mu,\Sigma,a,c\}$ denotes the Gaussian mean, covariance, opacity, or semantic feature, and $p$ is the top-1 semantic confidence. The parameter $\gamma\in(0,1)$ balances historical and newly observed information, and we set $\gamma<0.5$ to favor recent observations. After the update, new Gaussians that have not been assigned to any memory neighborhood are directly inserted into $\mathcal{M}$.
This strategy provides training-free, confidence-aware accumulation of sparse Gaussian representations over streaming observations.

\subsection{Training Objectives}\label{sec:loss}

We optimize both \modelname~and \newmodelname~using a composite objective
that combines semantic occupancy and geometric supervision. Following
EmbodiedOcc~\cite{embodiedocc}, we use focal loss
$L_{\mathrm{focal}}$, Lov\'asz-Softmax loss $L_{\mathrm{lov}}$, and geometric
and semantic scene-class affinity losses
$L^{\mathrm{geo}}_{\mathrm{scal}}$ and
$L^{\mathrm{sem}}_{\mathrm{scal}}$. We additionally apply a Huber loss
$L_{\mathrm{depth}}$ to the predicted depth, enabling end-to-end optimization
of the geometry-prior adaptation and occupancy prediction modules:
\begin{equation}\label{eq:loss}
\begin{split}
    \mathcal{L}
    ={}&
    L_{\mathrm{focal}}
    \bigl(
    Y_{\mathrm{pred}}^{\mathrm{fov}},
    Y_{\mathrm{gt}}^{\mathrm{fov}}
    \bigr)
    +
    L_{\mathrm{lov}}
    \bigl(
    Y_{\mathrm{pred}}^{\mathrm{fov}},
    Y_{\mathrm{gt}}^{\mathrm{fov}}
    \bigr)
    \\
    &+
    L^{\mathrm{geo}}_{\mathrm{scal}}
    \bigl(
    Y_{\mathrm{pred}}^{\mathrm{fov}},
    Y_{\mathrm{gt}}^{\mathrm{fov}}
    \bigr)
    +
    L^{\mathrm{sem}}_{\mathrm{scal}}
    \bigl(
    Y_{\mathrm{pred}}^{\mathrm{fov}},
    Y_{\mathrm{gt}}^{\mathrm{fov}}
    \bigr)
    +
    L_{\mathrm{depth}}.
\end{split}
\end{equation}

\section{\newmodelname}
\label{sec:gpoccpp}

\subsection{Overview}\label{sec:gpoccpp_overview}

\begin{figure*}[t]
  \centering
  \includegraphics[width=0.98\textwidth]{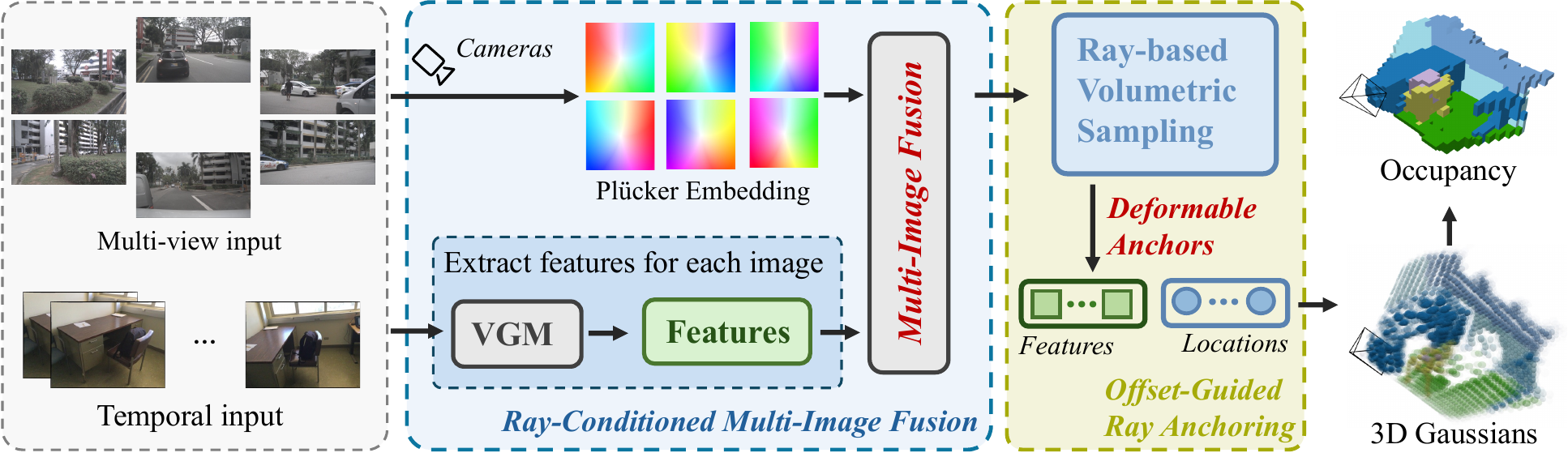}
  \caption{\textbf{Overview of \newmodelname.}
  Given temporal or multi-view images, a visual geometry model extracts
  per-image features and surface predictions, while Pl\"ucker embeddings
  encode the corresponding camera rays. Ray-Conditioned Multi-Image Fusion
  models cross-view or cross-temporal relationships in a unified feature
  space. The surface predictions define initial ray-based volumetric
  anchors, and the fused features condition Offset-Guided Ray Anchoring
  and Gaussian attribute prediction. The refined sparse Gaussians are
  subsequently mapped to semantic occupancy using the same probabilistic
  formulation as \modelname.}
  \label{fig:ppframwork}
\end{figure*}

Although \modelname~can use geometry priors produced from multiple images,
it processes their features largely in a frame-wise manner and constrains
Gaussian centers to their initial ray-sampled locations. As illustrated
in~\Cref{fig:ppframwork}, \newmodelname~addresses these limitations with
two extensions. Ray-Conditioned Multi-Image Fusion explicitly models
cross-view and cross-temporal feature relationships using camera-ray
geometry, while Offset-Guided Ray Anchoring refines the initial volumetric
anchors with learnable 3D residuals. The subsequent Gaussian prediction,
opacity pruning, occupancy aggregation, and incremental update follow the
same sparse formulation as \modelname.

\subsection{Ray-Conditioned Multi-Image Fusion}\label{sec:ccmif}

Multiple observations provide complementary geometric evidence in both embodied perception and autonomous driving. Temporal frames reveal scene regions over continuous exploration, while synchronized cameras cover different spatial viewpoints~\cite{chen2020visibility,lee2025visibility}. However, independently processing these images cannot explicitly model their geometric relationships.
Let the input image set be $\{\mathbf{I}_t\}_{t=1}^{T}$, where $T=1$ denotes the single-image case, and $T>1$ represents either temporally adjacent frames or synchronized multi-view images. For each image, the geometry-prior model extracts a feature map $\mathbf{F}_t\in\mathbb{R}^{H\times W\times C}$ together with its surface geometry prediction.

We introduce a lightweight interaction module that alternates between local and global attention~\cite{vggt}. Local attention first models intra-image spatial relationships:
\begin{equation}
    \bar{\mathbf{F}}_t
    =
    \operatorname{LocalAttn}(\mathbf{F}_t).
\end{equation}
The resulting tokens from all observations are then jointly processed by global attention:
\begin{equation}
    \tilde{\mathbf{F}}
    =
    \operatorname{GlobalAttn}
    \left(
    [
    \bar{\mathbf{F}}_1+\mathbf{E}^{\mathrm{cam}}_1;
    \ldots;
    \bar{\mathbf{F}}_T+\mathbf{E}^{\mathrm{cam}}_T
    ]
    \right),
\end{equation}
where $[\cdot;\cdot]$ denotes token concatenation and $\mathbf{E}^{\mathrm{cam}}_t$ encodes the camera-ray geometry of image $t$.

For temporal observations, global attention uses a causal mask~\cite{streamvggt}, allowing each frame to attend only to its current and preceding observations. During inference, historical tokens are retained in a cache~\cite{shazeer2019fast}, so only newly observed frames need to be processed.
To explicitly encode camera geometry, we represent each image ray using Pl\"ucker coordinates~\cite{plucker1865xvii,jiang2025rayzer}. For pixel $(u,v)$ in image $t$, let $\mathbf{o}_t$ be the camera center and $\mathbf{r}_{t,(u,v)}$ the normalized ray direction in the global coordinate system. The ray representation is
\begin{equation}
    \mathbf{p}_{t,(u,v)}
    =
    \left[
    \mathbf{r}_{t,(u,v)},
    \mathbf{o}_t\times\mathbf{r}_{t,(u,v)}
    \right],
\end{equation}
and is projected into the image feature space as
\begin{equation}
    \mathbf{E}^{\mathrm{cam}}_{t,(u,v)}
    =
    \operatorname{MLP}_{\mathrm{cam}}
    \left(\mathbf{p}_{t,(u,v)}\right).
\end{equation}
Adding these embeddings before global attention enables the model to jointly reason over appearance, camera configuration, and scene geometry.
The module outputs fused feature maps
$\{\tilde{\mathbf{F}}_t\}_{t=1}^{T}$ for subsequent anchor refinement
and Gaussian prediction, while explicitly modeling cross-view and
cross-temporal relationships.

\subsection{Offset-Guided Ray Anchoring}\label{sec:oara}

The ray-based anchors in \modelname~are constrained to fixed trajectories determined by the predicted depth and radial sampling offsets. Although this provides a strong geometric initialization, the optimal volumetric support may deviate from these initial positions for thin structures, irregular boundaries, partially occluded regions, or locally inaccurate geometry
predictions.
We therefore treat the initial ray samples as coarse anchors and predict a
3D residual for each anchor. For image $t$, the initial anchor associated
with pixel $(u,v)$ and sample index $k$ is
\begin{equation}
    \mathbf{x}^{0}_{t,u,v,k}
    =
    \mathbf{o}_t
    +
    \bigl(
    \mathbf{d}_{t,u,v}
    +
    \delta_{t,u,v,k}
    \bigr)
    \mathbf{r}_{t,u,v}.
\end{equation}

The anchor query combines the corresponding fused image feature with the learnable sampling-index embedding:
\begin{equation}
    \mathbf{q}_{t,u,v,k}
    =
    \tilde{\mathbf{F}}^{1/4}_{t,u,v}
    +
    \mathbf{E}_k.
\end{equation}
For a single-image input, $\tilde{\mathbf{F}}_t$ is replaced by the original geometry feature $\mathbf{F}_t$. A 3D residual offset is predicted as
\begin{equation}
    \Delta\mathbf{x}_{t,u,v,k}
    =
    \operatorname{MLP}_{\mathrm{off}}
    \left(\mathbf{q}_{t,u,v,k}\right),
\end{equation}
and the refined Gaussian center becomes
\begin{equation}
    \mathbf{x}_{t,u,v,k}
    =
    \mathbf{x}^{0}_{t,u,v,k}
    +
    \Delta\mathbf{x}_{t,u,v,k}.
\end{equation}
The remaining Gaussian attributes are predicted from the same anchor query:
\begin{equation}
    \{s_i,r_i,a_i,c_i\}
    =
    \operatorname{MLP}_{\mathrm{gs}}
    \left(\mathbf{q}_{t,u,v,k}\right).
\end{equation}

This design preserves the geometry-guided ray initialization while allowing
Gaussian centers to correct local depth errors and adapt to complex scene
structures. With multiple observations, the offsets are conditioned on fused
cross-image features; with a single observation, the module performs
monocular anchor refinement.

\section{Experiments}\label{sec:exp}

\begin{table*}[h] \small
    \caption{\textbf{Monocular prediction performance on the Occ-ScanNet dataset.}}
    \vspace{-1mm}
    \small
    \setlength{\tabcolsep}{0.008\textwidth}
    \captionsetup{font=scriptsize}
    \begin{center}
    \resizebox{0.98\linewidth}{!}{
    \begin{tabular}{l|c|c c c c c c c c c c c|c}
        \toprule
        Method
        & {IoU}
        & \rotatebox{90}{\parbox{1.5cm}{\textcolor{ceiling}{$\blacksquare$} ceiling}} 
        & \rotatebox{90}{\textcolor{floor}{$\blacksquare$} floor}
        & \rotatebox{90}{\textcolor{wall}{$\blacksquare$} wall} 
        & \rotatebox{90}{\textcolor{window}{$\blacksquare$} window} 
        & \rotatebox{90}{\textcolor{chair}{$\blacksquare$} chair} 
        & \rotatebox{90}{\textcolor{bed}{$\blacksquare$} bed} 
        & \rotatebox{90}{\textcolor{sofa}{$\blacksquare$} sofa} 
        & \rotatebox{90}{\textcolor{table}{$\blacksquare$} table} 
        & \rotatebox{90}{\textcolor{tvs}{$\blacksquare$} tvs} 
        & \rotatebox{90}{\textcolor{furniture}{$\blacksquare$} furniture} 
        & \rotatebox{90}{\textcolor{objects}{$\blacksquare$} objects} 
        & mIoU\\
        \midrule
        TPVFormer~\cite{Triformer} & 33.39 & 6.96 & 32.97 & 14.41 & 9.10 & 24.01 & 41.49 & 45.44 & 28.61 & 10.66 & 35.37 & 25.31 & 24.94 \\
        GaussianFormer~\cite{gaussianformer} & 40.91 & 20.70 & 42.00 & 23.40 & 17.40 & 27.0 & 44.30 & 44.80 & 32.70 & 15.30 & 36.70 & 25.00 & 29.93 \\
        MonoScene~\cite{monoscene} & 41.60 & 15.17 & 44.71 & 22.41 & 12.55 & 26.11 & 27.03 & 35.91 & 28.32 & 6.57 & 32.16 & 19.84 & 24.62 \\
        ISO~\cite{ISO} & 42.16 & 19.88 & 41.88 & 22.37 & 16.98 & 29.09 & 42.43 & 42.00 & 29.60 & 10.62 & 36.36 & 24.61 & 28.71 \\
        Surroundocc~\cite{surroundocc} & 42.52 & 18.90 & 49.30 & 24.80 & 18.00 & 26.80 & 42.00 & 44.10 & 32.90 & 18.60 & 36.80 & 26.90 & 30.83 \\
        EmbodiedOcc~\cite{embodiedocc} & 53.55 & 39.60 & 50.40 & 41.40 & 31.70 & 40.90 & 55.00 & 61.40 & 44.00 & 36.10 & 53.90 & 42.20 & 45.15 \\
        EmbodiedOcc++~\cite{embodiedocc++} & 54.90 & 36.40 & 53.10 & 41.80 & 34.40 & 42.90 & 57.30 & 64.10 & 45.20 & 34.80 & 54.20 & 44.10 & 46.20 \\
        RoboOcc~\cite{roboocc} & 56.48 & 45.36 & 53.49 & 44.35 & 34.81 & 43.38 & 56.93 & 63.35 & 46.35 & 36.12 & 55.48 & 44.78 & 47.67 \\
        \midrule
        \modelname~(DPT, Ours) & 56.96 & 51.42 & 50.35 & 46.97 & 41.84 & 46.98 & 60.39 & 66.16 & 50.51 & 47.97 & 58.88 & 49.23 & 51.88 \\
        \newmodelname~(DPT, Ours) & 57.68 & \underline{52.15} & 48.52 & 47.31 & 45.74 & 50.31 & 63.67 & \underline{69.77} & 52.87 & \underline{49.13} & 61.21 & 52.02 & 53.88 \\
        \gain{\textit{Improvement}} & \gain{+0.72} & \gain{+0.73} & \gain{-1.83} & \gain{+0.34} & \gain{+3.90} & \gain{+3.33} & \gain{+3.28} & \gain{+3.61} & \gain{+2.36} & \gain{+1.16} & \gain{+2.33} & \gain{+2.79} & \gain{+2.00} \\
        \midrule
        \modelname~(VGGT, Ours) & \underline{63.14} & 51.67 & \underline{59.93} & \textbf{52.07} & \underline{46.44} & \underline{51.35} & \underline{64.45} & 69.47 & \underline{54.30} & \textbf{51.76} & \underline{63.29} & \underline{53.36} & \underline{56.19} \\
        \newmodelname~(VGGT, Ours) & \textbf{63.72} & \textbf{56.62} & \textbf{61.40} & \underline{50.63} & \textbf{48.13} & \textbf{53.12} & \textbf{66.34} & \textbf{71.05} & \textbf{56.48} & 48.95 & \textbf{64.48} & \textbf{54.22} & \textbf{57.41} \\
        \gain{\textit{Improvement}} & \gain{+0.58} & \gain{+4.95} & \gain{+1.47} & \gain{-1.44} & \gain{+1.69} & \gain{+1.77} & \gain{+1.89} & \gain{+1.58} & \gain{+2.18} & \gain{-2.81} & \gain{+1.19} & \gain{+0.86} & \gain{+1.22} \\
        \bottomrule
    \end{tabular}
    }
    \end{center}
    \label{tab:main_mono}
 \end{table*}

\begin{table*}[h] 	\small
    \caption{\textbf{Embodied prediction performance on the EmbodiedOcc-ScanNet dataset.}}
    \vspace{-1mm}
    \setlength{\tabcolsep}{0.007\textwidth}
    \captionsetup{font=scriptsize}
    \begin{center}
    \resizebox{0.98\linewidth}{!}{
    \begin{tabular}{l|c|c c c c c c c c c c c|c}
        \toprule
        Method
        & {IoU}
        & \rotatebox{90}{\parbox{1.5cm}{\textcolor{ceiling}{$\blacksquare$} ceiling}} 
        & \rotatebox{90}{\textcolor{floor}{$\blacksquare$} floor}
        & \rotatebox{90}{\textcolor{wall}{$\blacksquare$} wall} 
        & \rotatebox{90}{\textcolor{window}{$\blacksquare$} window} 
        & \rotatebox{90}{\textcolor{chair}{$\blacksquare$} chair} 
        & \rotatebox{90}{\textcolor{bed}{$\blacksquare$} bed} 
        & \rotatebox{90}{\textcolor{sofa}{$\blacksquare$} sofa} 
        & \rotatebox{90}{\textcolor{table}{$\blacksquare$} table} 
        & \rotatebox{90}{\textcolor{tvs}{$\blacksquare$} tvs} 
        & \rotatebox{90}{\textcolor{furniture}{$\blacksquare$} furniture} 
        & \rotatebox{90}{\textcolor{objects}{$\blacksquare$} objects} 
        & mIoU\\
        \midrule
        TPVFormer~\cite{Triformer} & 35.88 & 1.62 & 30.54 & 12.03 & 13.22 & 35.47 & 51.39 & 49.79 & 25.63 & 3.60 & 43.15 & 16.23 & 25.70 \\
        SurroundOcc~\cite{surroundocc} & 37.04 & 12.70 & 31.80 & 22.50 & 22.00 & 29.90 & 44.70 & 36.50 & 24.60 & 11.50 & 34.40 & 18.20 & 26.27 \\
        GaussianFormer~\cite{gaussianformer} & 38.02 & 17.00 & 33.60 & 21.50 & 21.70 & 29.40 & 47.80 & 37.10 & 24.30 & 15.50 & 36.20 & 16.80 & 27.36 \\
        SplicingOcc~\cite{embodiedocc} & 49.01 & 31.60 & 38.80 & 35.50 & 36.30 & 47.10 & 54.50 & 57.20 & 34.40 & 32.50 & 51.20 & 29.10 & 40.74 \\
        EmbodiedOcc~\cite{embodiedocc} & 51.52 & 22.70 & 44.60 & 37.40 & 38.00 & 50.10 & 56.70 & 59.70 & 35.40 & 38.40 & 52.00 & 32.90 & 42.53 \\
        EmbodiedOcc++~\cite{embodiedocc++} & 52.20 & 27.90 & 43.90 & 38.70 & 40.60 & 49.00 & 57.90 & 59.20 & 36.80 & 37.80 & 53.50 & 34.10 & 43.60 \\
        RoboOcc~\cite{roboocc} & 53.30 & 21.94 & 44.57 & 39.53 & 38.48 & 51.28 & 57.04 & 63.09 & 36.70 & 43.05 & 54.42 & 34.38 & 44.05 \\
        \midrule
        \modelname~(DPT, Ours) & 56.39 & 40.80 & 48.78 & 45.62 & 43.26 & 50.08 & 63.97 & 67.72 & 48.36 & 48.77 & 60.46 & 45.63 & 51.22 \\
        \newmodelname~(DPT, Ours) & 60.89 & \textbf{48.94} & 50.07 & \underline{54.18} & \underline{52.73} & 52.70 & 67.07 & \underline{71.01} & 52.22 & 53.43 & \underline{65.85} & \underline{52.32} & \underline{56.41} \\
        \gain{\textit{Improvement}} & \gain{+4.50} & \gain{+8.14} & \gain{+1.29} & \gain{+8.56} & \gain{+9.47} & \gain{+2.62} & \gain{+3.10} & \gain{+3.29} & \gain{+3.86} & \gain{+4.66} & \gain{+5.39} & \gain{+6.69} & \gain{+5.19} \\
        \midrule
        \modelname~(VGGT, Ours) & \underline{61.41} & 42.61 & \underline{51.35} & 51.49 & 48.72 & \underline{54.32} & \underline{67.91} & 70.73 & \underline{52.94} & \underline{54.75} & 64.76 & 49.67 & 55.39 \\
        \newmodelname~(VGGT, Ours) & \textbf{62.50} & \underline{48.00} & \textbf{52.14} & \textbf{54.70} & \textbf{53.39} & \textbf{54.71} & \textbf{68.82} & \textbf{71.03} & \textbf{54.27} & \textbf{55.23} & \textbf{66.81} & \textbf{52.42} & \textbf{57.41} \\
        \gain{\textit{Improvement}} & \gain{+1.09} & \gain{+5.39} & \gain{+0.79} & \gain{+3.21} & \gain{+4.67} & \gain{+0.39} & \gain{+0.91} & \gain{+0.30} & \gain{+1.33} & \gain{+0.48} & \gain{+2.05} & \gain{+2.75} & \gain{+2.02} \\
        \bottomrule
    \end{tabular}
    }
    \end{center}
    \label{tab:main_online}
 \end{table*}

\begin{table*}[h] %
    \caption{\textbf{3D semantic occupancy prediction results on nuScenes.} 
    * means supervised by dense occupancy annotations as opposed to original LiDAR segmentation labels.}
    \small
    \setlength{\tabcolsep}{0.005\linewidth}  
    \renewcommand\arraystretch{1.05}
    \centering
    \resizebox{\textwidth}{!}{
    \begin{tabular}{l|c c | c c c c c c c c c c c c c c c c}
        \toprule
        Method
        & IoU
        & mIoU
        & \rotatebox{90}{\textcolor{nbarrier}{$\blacksquare$} barrier}
        & \rotatebox{90}{\textcolor{nbicycle}{$\blacksquare$} bicycle}
        & \rotatebox{90}{\textcolor{nbus}{$\blacksquare$} bus}
        & \rotatebox{90}{\textcolor{ncar}{$\blacksquare$} car}
        & \rotatebox{90}{\textcolor{nconstruct}{$\blacksquare$} const. veh.}
        & \rotatebox{90}{\textcolor{nmotor}{$\blacksquare$} motorcycle}
        & \rotatebox{90}{\textcolor{npedestrian}{$\blacksquare$} pedestrian}
        & \rotatebox{90}{\textcolor{ntraffic}{$\blacksquare$} traffic cone}
        & \rotatebox{90}{\textcolor{ntrailer}{$\blacksquare$} trailer}
        & \rotatebox{90}{\textcolor{ntruck}{$\blacksquare$} truck}
        & \rotatebox{90}{\textcolor{ndriveable}{$\blacksquare$} drive. suf.}
        & \rotatebox{90}{\textcolor{nother}{$\blacksquare$} other flat}
        & \rotatebox{90}{\textcolor{nsidewalk}{$\blacksquare$} sidewalk}
        & \rotatebox{90}{\textcolor{nterrain}{$\blacksquare$} terrain}
        & \rotatebox{90}{\textcolor{nmanmade}{$\blacksquare$} manmade}
        & \rotatebox{90}{\textcolor{nvegetation}{$\blacksquare$} vegetation}
        \\
        \midrule
        MonoScene~\cite{monoscene} & 23.96 & 7.31 & 4.03 &	0.35& 8.00& 8.04&	2.90& 0.28& 1.16&	0.67&	4.01& 4.35&	27.72&	5.20& 15.13&	11.29&	9.03&	14.86 \\
        
        Atlas~\cite{atlas} & 28.66 & 15.00 & 10.64&	5.68&	19.66& 24.94& 8.90&	8.84&	6.47& 3.28&	10.42&	16.21&	34.86&	15.46&	21.89&	20.95&	11.21&	20.54 \\
        
        BEVFormer~\cite{bevformer} & 30.50 & 16.75 & 14.22 &	6.58 & 23.46 & 28.28& 8.66 &10.77& 6.64& 4.05& 11.20&	17.78 & 37.28 & 18.00 & 22.88 & 22.17 & {13.80} & 22.21 \\
        
        TPVFormer~\cite{Triformer} & 11.51 & 11.66 & 16.14&	7.17& 22.63	& 17.13 & 8.83 & 11.39 & 10.46 & 8.23&	9.43 & 17.02 & 8.07 & 13.64 & 13.85 & 10.34 & 4.90 & 7.37\\
        
        TPVFormer*~\cite{Triformer}  & {30.86} & 17.10 & 15.96&	 5.31& 23.86	& 27.32 & 9.79 & 8.74 & 7.09 & 5.20& 10.97 & 19.22 & {38.87} & {21.25} & {24.26} & {23.15} & 11.73 & 20.81\\

        OccFormer~\cite{occformer} & {31.39} & {19.03} & {18.65} & {10.41} & {23.92} & \underline{30.29} & {10.31} & {14.19} & \underline{13.59} & {10.13} & {12.49} & {20.77} & {38.78} & 19.79 & 24.19 & 22.21 & {13.48} & {21.35}\\

        SurroundOcc~\cite{surroundocc} & {31.49} & \underline{20.30} & \textbf{20.59} & {11.68} & \textbf{28.06} & \textbf{30.86} & {10.70} & {15.14} & \textbf{14.09} & \textbf{12.06} & \textbf{14.38} & \textbf{22.26} & 37.29 & {23.70} & {24.49} & {22.77} & \underline{14.89} & {21.86}  \\

        GaussianFormer~\cite{gaussianformer} & 29.83 & {19.10} & {19.52} & {11.26} & {26.11} & {29.78} & {10.47} & {13.83} & {12.58} & {8.67} & {12.74} & \underline{21.57} & {39.63} & {23.28} & {24.46} & {22.99} & 9.59 & 19.12 \\

        GaussianFormer-2~\cite{gaussianformer2} & 30.56 & {20.02} & \underline{20.15} & \underline{12.99} & \underline{27.61} & {30.23} & {11.19} & \underline{15.31} & {12.64} & {9.63} & {13.31} & \textbf{22.26} & {39.68} & {23.47} & {25.62} & {23.20} & 12.25 & 20.73 \\

        QuadricFormer~\cite{zuo2025quadricformer} & 31.22 & 20.12 & 19.58 & \textbf{13.11} & 27.27 & 29.64 & 11.25 & \textbf{16.26} & 12.65 & 9.15 & 12.51 & 21.24 & \underline{40.20} & \underline{24.34} & \underline{25.69} & 24.24 & 12.95 & 21.86 \\

        \midrule
        \modelname~(Ours) & \underline{31.92} & 19.83 & 18.87 & 10.97 & 23.40 & 27.49 & \underline{12.26} & 13.35 & 12.42 & \underline{11.30} & 13.52 & 19.95 & 39.72 & 24.19 & 25.64 & \underline{25.08} & 14.57 & \underline{24.59} \\

        \newmodelname~(Ours) & \textbf{33.28} & \textbf{20.78} & 19.72 & 11.59 & 26.36 & 28.31 & \textbf{12.69} & 14.76 & 12.22 & 10.67 & \underline{13.62} & 20.04 & \textbf{42.07} & \textbf{26.93} & \textbf{27.30} & \textbf{26.34} & \textbf{15.27} & \textbf{24.61} \\
        \gain{\textit{Improvement}} & \gain{+1.36} & \gain{+0.95} & \gain{+0.85} & \gain{+0.62} & \gain{+2.96} & \gain{+0.82} & \gain{+0.43} & \gain{+1.41} & \gain{-0.20} & \gain{-0.63} & \gain{+0.10} & \gain{+0.09} & \gain{+2.35} & \gain{+2.74} & \gain{+1.66} & \gain{+1.26} & \gain{+0.70} & \gain{+0.02} \\

        \bottomrule
    \end{tabular}}
    \label{tab:nuscenes_results}
\end{table*}

In this section, we comprehensively evaluate \modelname~and its extended version, \newmodelname, across indoor monocular occupancy prediction, embodied scene-level occupancy prediction, and outdoor multi-view occupancy prediction. We compare them with representative state-of-the-art approaches on Occ-ScanNet, EmbodiedOcc-ScanNet, and nuScenes, and conduct ablation studies to validate the key designs of \modelname~and the additional components in \newmodelname. We further present model profiling and qualitative analyses to examine accuracy--efficiency trade-offs and reconstruction quality.

\subsection{Datasets and Metrics}

\noindent\textbf{Occ-ScanNet}~\cite{ISO} is a large-scale benchmark for monocular indoor occupancy prediction, containing 45,755 training samples and 19,764 testing samples. The dataset covers diverse scenes and viewpoints, and provides voxelized frames in $60 \times 60 \times 36$ grids, corresponding to a $4.8 \text{m} \times 4.8 \text{m} \times 2.88 \text{m}$ volume in front of the camera. Each voxel is annotated with 12 semantic classes, including 11 valid categories (ceiling, floor, wall, window, chair, bed, sofa, table, TV, furniture, objects) and one class for empty space.

\noindent\textbf{EmbodiedOcc-ScanNet}~\cite{embodiedocc} is a reorganized version of Occ-ScanNet, consisting of 537 training scenes and 137 validation scenes. Each scene contains 30 posed frames, and the global occupancy resolution of a scene is defined as $\frac{l_x \times l_y \times l_z}{(0.08\text{m})^3}$, where $l_x \times l_y \times l_z$ denotes the spatial range of the scene in world coordinates.

\noindent\textbf{nuScenes}~\cite{nuscenes} is a large-scale benchmark for autonomous driving, containing 1000 surround-view driving scenes collected in Boston and Singapore. Following the official split, the dataset is divided into 700, 150, and 150 scenes for training, validation, and testing, respectively. Each scene is 20 seconds long and annotated at 2Hz, with synchronized sensor data from 6 cameras, 1 LiDAR, 5 radars, and 1 IMU. Following prior works~\cite{surroundocc,gaussianformer2,zuo2025quadricformer}, we use the 3D semantic occupancy annotations provided by SurroundOcc~\cite{surroundocc} for supervision and evaluation. The annotated occupancy space covers the range of $[-50, 50]$ m, $[-50, 50]$ m, and $[-5, 3]$ m along the $x$, $y$, and $z$ axes, respectively. Each voxel has a side length of 0.5 m and is assigned one of 18 classes, including 16 semantic classes, one empty class, and one noise class.

\noindent\textbf{Evaluation Metrics.}
We follow prior work and adopt mIoU and IoU as evaluation metrics. Specifically, for Occ-ScanNet~\cite{ISO,embodiedocc}, we compute IoU between predictions and ground truth within the camera frustum of each frame. For EmbodiedOcc-ScanNet~\cite{embodiedocc}, we evaluate global occupancy by computing IoU at the scene level, where the entire reconstructed scene is considered. For nuScenes~\cite{nuscenes}, we follow the standard protocol in prior occupancy prediction works~\cite{surroundocc,gaussianformer2,zuo2025quadricformer} and report mIoU over semantic occupancy predictions in the predefined 3D voxel space.

\subsection{Implementation Details}

\noindent\textbf{Indoor benchmarks.}
For the Occ-ScanNet benchmarks, we adopt a unified training strategy for all models. We use the AdamW optimizer~\cite{adamw} with a weight decay of 0.01. The learning rate is linearly warmed up during the first 1000 iterations to a maximum value of $2\times10^{-4}$ and then decayed with a cosine schedule. The models are trained for 10 epochs with a total batch size of 8 on 4 NVIDIA A800 GPUs. Input images are resized such that the longer side is 518 pixels, following VGGT~\cite{vggt}. We apply gradient clipping with a maximum norm of 1.0. Unless otherwise specified, we set $K=16$ for ray-based volumetric sampling, and $\tau=0.01$ for opacity-based pruning.

\noindent\textbf{Temporal training.}
For temporal indoor experiments, we adopt a two-stage training strategy. We first train a monocular \newmodelname~model equipped with the proposed temporal modules, where only the current frame is used as input. This stage provides a strong initialization while keeping the optimization close to the single-frame setting. We then use the resulting model as the pretrained model and continue training with temporal inputs. During temporal training, frames are processed in chronological order, and features from previous frames are cached and fed to the current frame, enabling causal multi-frame reasoning without accessing future observations.

\noindent\textbf{Outdoor benchmark.}
For nuScenes, we adopt a separate training setup tailored to outdoor autonomous driving scenes following previous works~\cite{gaussianformer2,zuo2025quadricformer}. We use DepthAnything-V2-Base~\cite{depthanythingv2} as the backbone, since its latency is close to current state-of-the-art occupancy methods. We use the AdamW optimizer with a base learning rate of $4\times10^{-4}$ and a weight decay of 0.01. The model is trained for 20 epochs with a total batch size of 8, and gradient clipping with a maximum norm of 1.0 is also applied. We use six surround-view camera images as input, and follow the standard nuScenes occupancy setting with a 3D range of $[-50,50]\times[-50,50]\times[-5,3]$ meters and a voxel size of 0.5 meters. Input images are resized such that the longer side is 518 pixels.

\begin{figure*}[t]
  \centering
  \includegraphics[width=\textwidth]{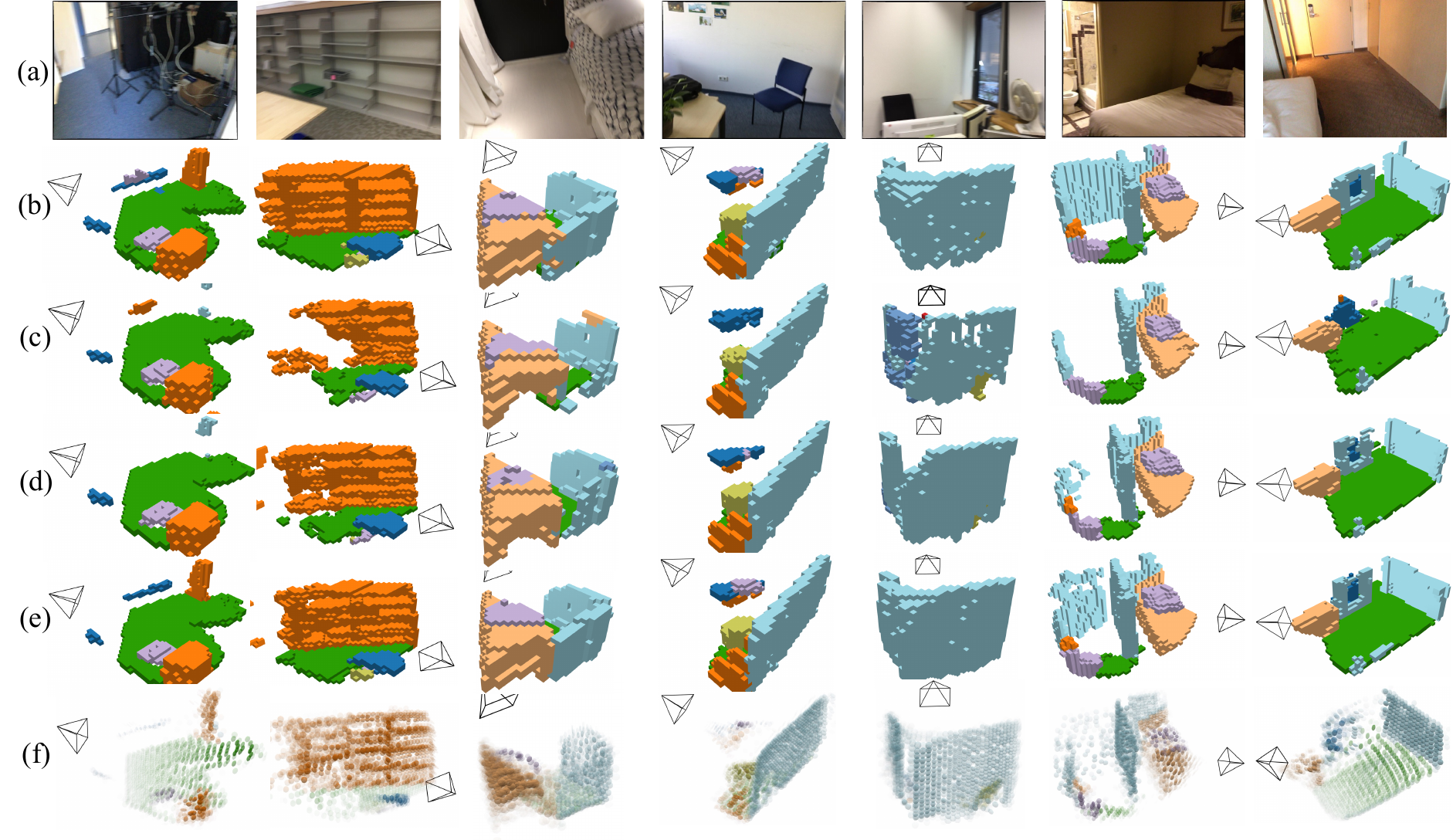}
  \caption{\textbf{Qualitative comparison on monocular occupancy prediction.}
 (a) shows the input RGB images, (b) the ground-truth occupancy, (c) the predictions of EmbodiedOcc~\cite{embodiedocc}, (d) the predictions of \modelname, (e) the predictions of \newmodelname, and (f) the Gaussian primitives predicted by \newmodelname. Compared to EmbodiedOcc and \modelname, \newmodelname~produces more accurate and complete reconstructions, while the Gaussian representation provides interpretable intermediate geometry.}
  \vspace{5mm}
  \label{fig:mono_vis}
\end{figure*}

\subsection{Occupancy Prediction Results}

\noindent\textbf{Results on Occ-ScanNet.}
As shown in~\Cref{tab:main_mono}, bold numbers denote the best result and underlined numbers denote the second best. Overall, \newmodelname~(VGGT) achieves the best result with 63.72 IoU and 57.41 mIoU, surpassing EmbodiedOcc and EmbodiedOcc++ by +12.26 and +11.21 mIoU, respectively.
Using DepthAnything as the geometry prior, same as~\cite{embodiedocc,ISO}, Ours-DPT model already delivers consistent improvements across classes over previous methods and improves mIoU from 46.20 (EmbodiedOcc++) to 51.88. The \newmodelname~extension further raises DPT-based mIoU to 53.88. Furthermore, replacing DepthAnything with the stronger VGGT prior yields additional gains, showing that our framework generalizes well across different geometry priors and can fully benefit from more powerful foundation models.

\noindent\textbf{Results on EmbodiedOcc-ScanNet.} 
\Cref{tab:main_online} reports the embodied prediction performance on the EmbodiedOcc-ScanNet benchmark. 
Our base \modelname~already achieves strong performance in the embodied setting. With DepthAnything as the geometry prior, \modelname~(DPT) attains 56.39 IoU and 51.22 mIoU, outperforming EmbodiedOcc++~\cite{embodiedocc++} by +4.19 IoU and +7.62 mIoU. Replacing DepthAnything with the stronger VGGT prior further improves the result to 61.41 IoU and 55.39 mIoU, showing that \modelname~can effectively exploit stronger geometry priors.
With the proposed \newmodelname~extensions, the semantic occupancy quality is further improved. \newmodelname~(DPT) reaches 60.89 IoU and 56.41 mIoU, improving over \modelname~(DPT) by +4.50 IoU and +5.19 mIoU. \newmodelname~(VGGT) obtains the best results with 62.50 IoU and 57.41 mIoU, which is +10.30 IoU and +13.81 mIoU higher than EmbodiedOcc++ and +1.09 IoU and +2.02 mIoU higher than \modelname~(VGGT). These gains indicate that Ray-Conditioned Multi-Image Fusion and Offset-Guided Ray Anchoring improve both geometric completeness and semantic consistency in the global embodied scene. These results show that the proposed extensions are particularly beneficial for embodied occupancy prediction, where sequential observations must be fused into a coherent scene-level representation.

\noindent\textbf{Results on nuScenes.}
\Cref{tab:nuscenes_results} reports the semantic occupancy prediction results on nuScenes.
Different from indoor embodied benchmarks, nuScenes focuses on large-scale outdoor driving scenes with surround-view camera inputs, long-range spatial layouts, dynamic traffic participants, and complex road structures. This setting introduces a distinct type of challenge, requiring the model to recover both global scene layout and fine-grained semantic geometry from multi-view observations.
\modelname~achieves competitive performance with 31.92 IoU and 19.83 mIoU. \newmodelname~further improves the performance to 33.28 IoU and 20.78 mIoU, bringing gains of +1.36 IoU and +0.95 mIoU over \modelname.
Compared with prior Gaussian-based methods, \newmodelname~achieves stronger overall accuracy while preserving the efficiency and compactness of the sparse representation.
The improvements are also reflected in several structure-sensitive categories, such as driveable surface, other flat, sidewalk, indicating that \newmodelname~recovers more complete scene layouts and richer semantic details in outdoor driving environments.
These results show that our method can effectively extend from indoor monocular occupancy prediction to multi-camera semantic occupancy reconstruction in large-scale autonomous driving scenes.

\subsection{\modelname~ablations}
We first analyze the key design choices in \modelname, including ray sampling and opacity pruning on Occ-ScanNet. 
Detailed sensitivity analyses of the sampling number $K$ and opacity threshold $\tau$, together with ablations on the incremental update strategy, are provided in the Appendix.

\begin{table}[h]
\small
\centering
\vspace{-1mm}
\caption{\textbf{Ablation of volumetric ray sampling and opacity pruning on Occ-ScanNet.}
}
\label{tab:abl_sampling_pruning}
\begin{tabular}{l|cc}
\toprule
Setting & mIoU & IoU \\
\midrule
w/o volumetric ray sampling & 47.88 & 53.10 \\
w/o opacity pruning & \textbf{56.81} & \textbf{63.53} \\
\modelname & 56.19 & 63.14 \\
\bottomrule
\end{tabular}
\end{table}

\noindent\textbf{Effect of volumetric ray sampling and opacity pruning.}
As shown in~\Cref{tab:abl_sampling_pruning}, restricting each camera ray to a single surface sample substantially reduces the mIoU from 56.19 to 47.88. This result confirms that extending surface-centric geometry priors along camera rays is essential for recovering volumetric object support beyond the visible surfaces. Removing opacity pruning yields only modest improvements of 0.62 mIoU and 0.39 IoU, while retaining all low-opacity Gaussian primitives and consequently increasing representation redundancy. The full model therefore adopts opacity pruning to achieve a more favorable trade-off between occupancy accuracy and representation compactness.

\begin{figure*}[t]
  \centering
  \includegraphics[width=\textwidth]{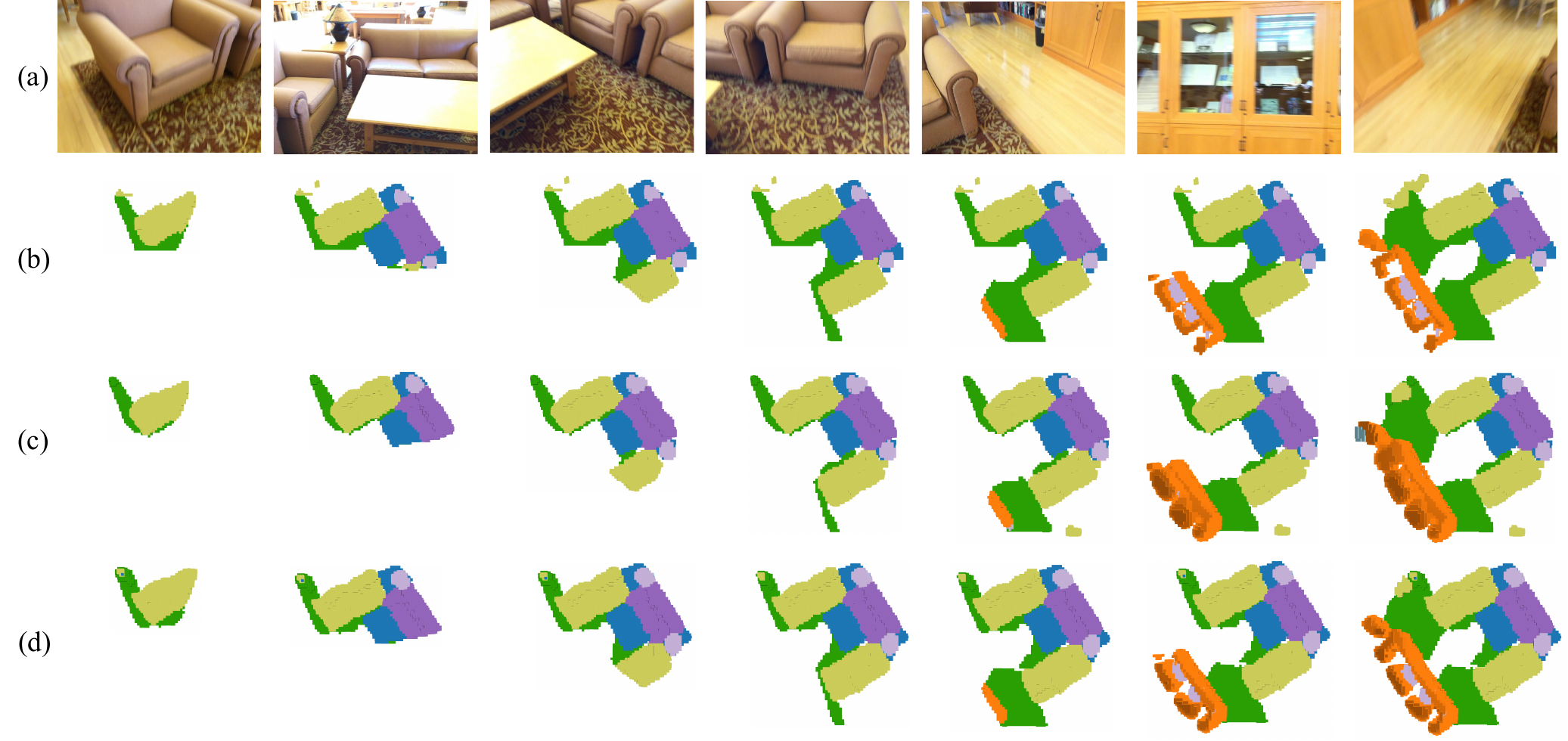}
  \caption{\textbf{Qualitative results on streaming inputs.}
  (a) shows the input RGB images, (b) the ground-truth occupancy, (c) the predictions of \modelname, and (d) the predictions of \newmodelname. Our incremental update strategy progressively integrates information from sequential frames, and \newmodelname~produces more complete and consistent predictions.}
  \label{fig:emb_vis}
\end{figure*}

\subsection{\newmodelname~ablations}
We evaluate the two key components of \newmodelname, namely Ray-Conditioned Multi-Image Fusion and Offset-Guided Ray Anchoring, on both indoor embodied benchmarks and outdoor autonomous driving datasets.

\begin{table}[h]\small
\caption{\textbf{Comparison between \modelname~and \newmodelname~on monocular and embodied occupancy prediction.}
DPT and VGGT refer to~\cite{depthanythingv2} and~\cite{vggt}, respectively. Monocular results are evaluated on Occ-ScanNet, while embodied results are evaluated on EmbodiedOcc-ScanNet.}
\vspace{-1mm}
\centering
\resizebox{\linewidth}{!}{
    \begin{tabular}{llcc|cc}
    \toprule
    \multirow{2}{*}{Geometry prior} & \multirow{2}{*}{Setting}
    & \multicolumn{2}{c|}{Monocular}
    & \multicolumn{2}{c}{Embodied} \\
    & & IoU & mIoU & IoU & mIoU \\
    \midrule
    DPT & \modelname & 56.96 & 51.88 & 56.39 & 51.22 \\
    DPT & + Multi-Image Fusion & 57.11 & 52.26 & 58.72 & 54.48 \\
    DPT & \newmodelname & 57.68 & 53.88 & 60.89 & 56.41 \\
    \midrule
    VGGT & \modelname & 63.14 & 56.19 & 61.41 & 55.39 \\
    VGGT & + Multi-Image Fusion & 63.29 & 56.73 & 61.98 & 56.43 \\
    VGGT & \newmodelname & 63.72 & 57.41 & 62.50 & 57.41 \\
    \bottomrule
    \end{tabular}
}
\label{tab:component_abl}
\end{table}

\noindent\textbf{Effect of \newmodelname~components on Occ-ScanNet.}
\Cref{tab:component_abl} compares \modelname~and \newmodelname~under monocular and embodied settings. On Occ-ScanNet, Ray-Conditioned Multi-Image Fusion consistently improves performance, increasing mIoU from 51.88 to 52.26 with DPT and from 56.19 to 56.73 with VGGT. Incorporating Offset-Guided Ray Anchoring further improves performance to 53.88 and 57.41 mIoU, respectively. These results indicate that both components benefit even single-image occupancy prediction by enhancing cross-token interaction and improving geometric alignment.

\noindent\textbf{Effect of \newmodelname~components on EmbodiedOcc-ScanNet.}
The improvements are more pronounced in embodied settings as shown in~\Cref{tab:component_abl}. With DPT, Ray-Conditioned Multi-Image Fusion improves IoU/mIoU from 56.39/51.22 to 58.72/54.48, and the full \newmodelname~further boosts performance to 60.89/56.41. With VGGT, the corresponding results improve from 61.41/55.39 to 61.98/56.43 and then to 62.50/57.41. This demonstrates that explicit modeling of inter-image relationships is particularly beneficial for embodied perception, where sequential observations must be aggregated into a coherent scene representation.

\begin{table}[h]\small
\caption{\textbf{\newmodelname~component ablation on nuScenes.}}
\centering
\resizebox{0.65\linewidth}{!}{
    \begin{tabular}{lcc}
    \toprule
    Setting & IoU & mIoU \\
    \midrule
    \modelname & 31.92 & 19.83 \\
    + Multi-Image Fusion & 32.04 & 20.03 \\
    \newmodelname & 33.28 & 20.78 \\
    \bottomrule
    \end{tabular}
}
\label{tab:nusc_component_abl}
\end{table}

\noindent\textbf{Effect of \newmodelname~components on nuScenes.}
\Cref{tab:nusc_component_abl} shows the corresponding \newmodelname~component ablation on nuScenes. Ray-Conditioned Multi-Image Fusion improves the baseline from 31.92 IoU and 19.83 mIoU to 32.04 IoU and 20.03 mIoU. The full \newmodelname~with both Ray-Conditioned Multi-Image Fusion and Offset-Guided Ray Anchoring further improves the result to 33.28 IoU and 20.78 mIoU, showing that the two modules provide complementary benefits in large-scale outdoor scenes.

\begin{figure*}[t]
  \centering
  \includegraphics[width=\textwidth]{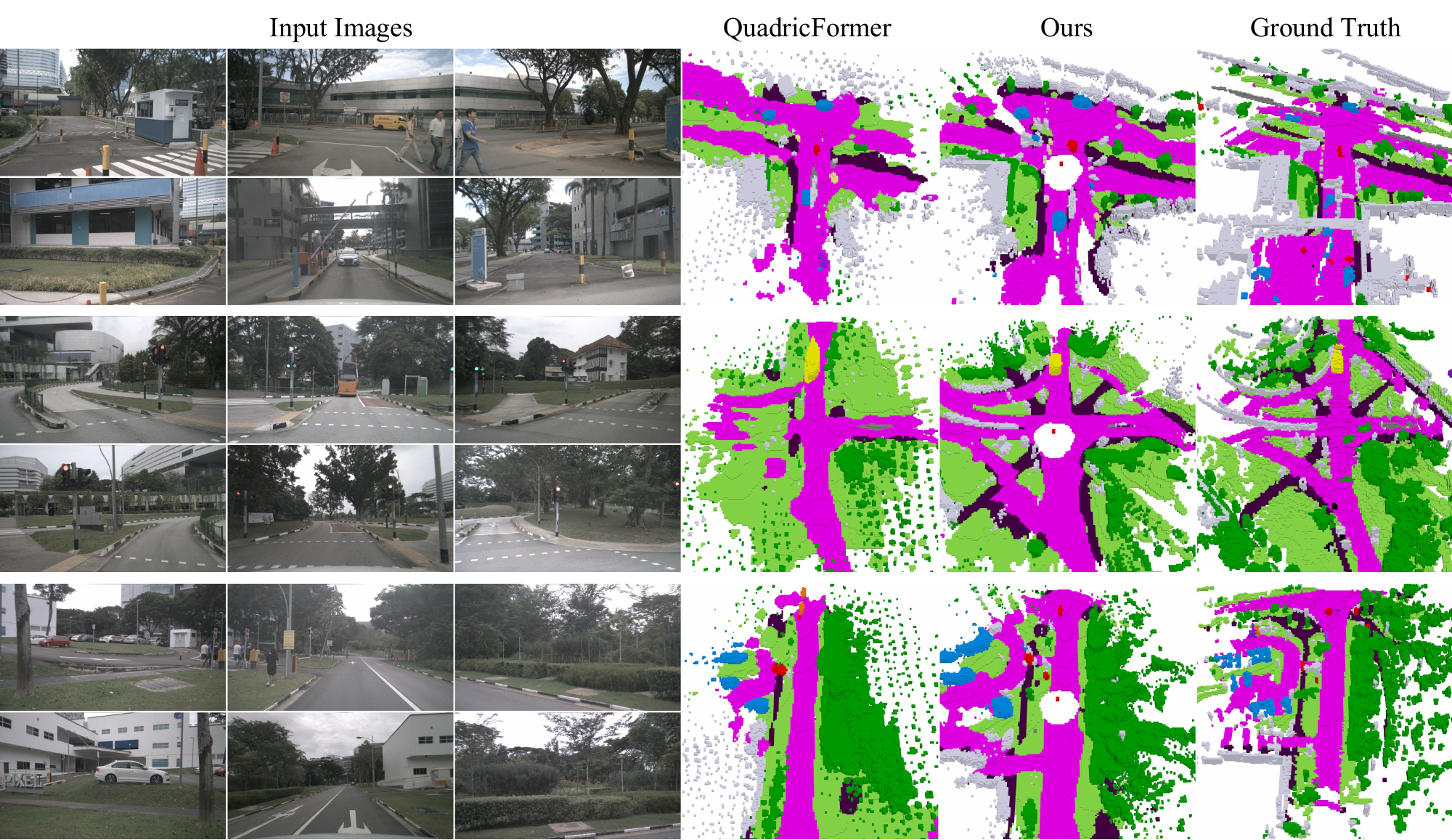}
  \caption{\textbf{Qualitative results on nuScenes.}
  We visualize surround-view camera inputs, the semantic occupancy predicted by \newmodelname, and the corresponding ground-truth occupancy in outdoor driving scenes.}
  \label{fig:nusc_vis}
\end{figure*}

\begin{table}[h]
\scriptsize
\centering
\caption{\textbf{Model profile on Occ-ScanNet.}}
\vspace{-1mm}
\label{tab:profile_occ_scannet}
\resizebox{\linewidth}{!}{
\begin{tabular}{lcccc}
\toprule
Model & IoU & mIoU & FPS & \#Params \\
\midrule
ISO~\cite{ISO} & 42.16 & 28.71 & 3.63 & 303.05M \\
EmbodiedOcc~\cite{embodiedocc} & 53.55 & 45.15 & 10.66 & 231.45M \\
\midrule
\modelname~(DPT, Ours) & 56.96 & 51.88 & 28.22 & 97.95M \\
\modelname~(VGGT, Ours) & 63.14 & 56.19 & 5.26 & 942.31M \\
\midrule
\newmodelname~(DPT, Ours) & 57.68 & 53.88 & 23.85 & 131.49M \\
\newmodelname~(VGGT, Ours) & 63.72 & 57.41 & 5.13 & 954.01M \\
\bottomrule
\end{tabular}
}
\end{table}

\noindent\textbf{Model profile.}
\Cref{tab:profile_occ_scannet,tab:profile_embodiedocc_scannet,tab:profile_nuscenes} summarize the accuracy, inference speed, and model size across indoor, embodied, and outdoor settings.
DPT and VGGT refer to~\cite{depthanythingv2} and~\cite{vggt}, respectively. FPS values are measured on the same NVIDIA A800 GPU and averaged over 1000 runs after 100 warm-up iterations.
On Occ-ScanNet, \modelname~(DPT, Ours) achieves strong efficiency, running at 28.22 FPS with only 97.95M parameters, which is nearly $3\times$ faster and less than half the size of EmbodiedOcc.
\newmodelname~(DPT, Ours) still maintains 23.85 FPS with moderate parameter overhead, while VGGT variants further improve accuracy with a larger backbone.

\begin{table}[h]
\scriptsize
\centering
\caption{\textbf{Model profile on EmbodiedOcc-ScanNet.}}
\vspace{-1mm}
\label{tab:profile_embodiedocc_scannet}
\resizebox{\linewidth}{!}{
\begin{tabular}{lcccc}
\toprule
Model & IoU & mIoU & FPS & \#Params \\
\midrule
EmbodiedOcc~\cite{embodiedocc} & 51.52 & 42.53 & 10.41 & 231.45M \\
\midrule
\modelname~(DPT, Ours) & 56.39 & 51.22 & 27.36 & 97.95M \\
\modelname~(VGGT, Ours) & 61.41 & 55.39 & 5.02 & 942.31M \\
\midrule
\newmodelname~(DPT, Ours) & 60.89 & 56.41 & 19.58 & 131.49M \\
\newmodelname~(VGGT, Ours) & 62.50 & 57.41 & 4.31 & 954.01M \\
\bottomrule
\end{tabular}
}
\end{table}

On EmbodiedOcc-ScanNet, DPT-based variants remain compact and efficient, while VGGT-based variants provide higher accuracy.
Compared with EmbodiedOcc, \modelname~(DPT, Ours) improves mIoU by 8.69 points and runs $2.63\times$ faster with less than half of the parameters.
\newmodelname~further improves both backbones, showing that its causal temporal modeling brings consistent gains with limited additional cost.

\begin{table}[h]
\scriptsize
\centering
\caption{\textbf{Model profile on nuScenes.}}
\vspace{-1mm}
\label{tab:profile_nuscenes}
\resizebox{\linewidth}{!}{
\begin{tabular}{lcccc}
\toprule
Model & IoU & mIoU & FPS & \#Params \\
\midrule
GaussianFormer-2~\cite{gaussianformer2} & 30.56 & 20.02 & 1.85 & 71.57M \\
QuadricFormer~\cite{zuo2025quadricformer} & 31.22 & 20.12 & 4.58 & 71.46M \\
\midrule
\modelname~(Ours) & 31.92 & 19.83 & 8.51 & 150.19M \\
\newmodelname~(Ours) & 33.28 & 20.78 & 8.02 & 161.45M \\
\bottomrule
\end{tabular}
}
\end{table}

On nuScenes, \newmodelname~(Ours) runs at 8.02 FPS, faster than GaussianFormer-2 and QuadricFormer, while achieving the best overall accuracy.
These results demonstrate a favorable trade-off between accuracy and efficiency across both indoor embodied perception and outdoor autonomous driving scenarios.
For FPS interpretation on nuScenes, we note that our method uses images resized to a maximum side length of 518, while the compared methods follow the official $900 \times 1600$ input resolution.

\subsection{Qualitative Results}

\paragraph{Indoor scenes}
\Cref{fig:mono_vis} shows qualitative comparisons on indoor monocular occupancy prediction. \newmodelname~produces more complete and geometrically consistent occupancy predictions than EmbodiedOcc~\cite{embodiedocc} and \modelname, especially for large structures such as walls, floors, furniture, and shelves.
The predicted Gaussian primitives provide an interpretable intermediate geometry, showing that the model learns compact 3D scene structure before decoding it into voxel occupancy.
\Cref{fig:emb_vis} presents qualitative results on streaming indoor inputs. With sequential observations, our incremental update strategy progressively integrates newly observed regions and refines incomplete predictions, leading to more complete and consistent occupancy reconstruction.

\paragraph{Outdoor scenes}
\Cref{fig:nusc_vis} shows qualitative results on outdoor driving scenes from nuScenes. Given surround-view camera images, \newmodelname~predicts semantic occupancy that is well aligned with the ground truth and preserves more complete road layouts, object-level structures, and fine-grained scene details than QuadricFormer. These results demonstrate that \newmodelname~can recover coherent semantic geometry in complex driving scenarios with long-range spatial structures.

More visualizations are provided in the Appendix.

\section{Limitations}

Although \newmodelname~shows strong generality across indoor and outdoor benchmarks, several limitations remain. First, the method can still be less accurate on large textureless or flat regions such as floors and roads, where geometry priors may provide weak local evidence and the sparse Gaussian representation may under-cover broad surfaces. Second, the incremental update strategy used for streaming embodied inputs continuously inserts new Gaussian primitives, which may increase memory and computation for long sequences. Third, the current formulation mainly focuses on static scenes; explicitly modeling dynamic objects and maintaining long-term maps remain important directions for future work.

\section{Conclusion}

We presented \newmodelname, a general and flexible framework for geometry-prior-based occupancy prediction across both embodied and autonomous driving scenarios. Building upon \modelname, the proposed framework introduces Ray-Conditioned Multi-Image Fusion to explicitly model relations across temporal and multi-view observations with camera-ray geometry. In addition, Offset-Guided Ray Anchoring replaces the original fixed ray sampling scheme by directly predicting 3D residuals for sampled anchors, better aligning Gaussian centers with complex structures. Together with sparse Gaussian representation and incremental update for streaming inputs, our approach provides an effective and scalable solution for fine-grained 3D occupancy prediction.
Extensive experiments on Occ-ScanNet, EmbodiedOcc-ScanNet, and nuScenes demonstrate that \newmodelname~achieves strong performance across diverse indoor and outdoor benchmarks, while retaining favorable efficiency and generality across different geometry priors. We hope this work can serve as a step toward more unified 3D scene understanding systems that effectively integrate strong visual geometry priors for a broad range of embodied and autonomous perception tasks.

\section*{Acknowledgment}

This work was supported in part by the National Natural Science Foundation
of China under Grant 62573370 and in part by the Education Department of
Guangdong Province under Grant 2025ZDZX3051.

\FloatBarrier

{
    \bibliographystyle{IEEEtran}
    \bibliography{main}

@String(CVPR= {IEEE Conf. Comput. Vis. Pattern Recog.})

@String(ICCV= {Int. Conf. Comput. Vis.})

@String(ECCV= {Eur. Conf. Comput. Vis.})

@String(CVPR  = {CVPR})

@String(ICCV  = {ICCV})

@String(ECCV  = {ECCV})

@inproceedings{embodiedocc,
  title={Embodiedocc: Embodied 3d occupancy prediction for vision-based online scene understanding},
  author={Wu, Yuqi and Zheng, Wenzhao and Zuo, Sicheng and Huang, Yuanhui and Zhou, Jie and Lu, Jiwen},
  booktitle={Proceedings of the IEEE/CVF International Conference on Computer Vision},
  pages={26360--26370},
  year={2025}
}

@article{embodiedocc++,
  title={Embodiedocc++: Boosting embodied 3d occupancy prediction with plane regularization and uncertainty sampler},
  author={Wang, Hao and Wei, Xiaobao and Zhang, Xiaoan and Li, Jianing and Bai, Chengyu and Li, Ying and Lu, Ming and Zheng, Wenzhao and Zhang, Shanghang},
  journal={arXiv preprint arXiv:2504.09540},
  year={2025}
}

@inproceedings{gaussianformer,
  title={Gaussianformer: Scene as gaussians for vision-based 3d semantic occupancy prediction},
  author={Huang, Yuanhui and Zheng, Wenzhao and Zhang, Yunpeng and Zhou, Jie and Lu, Jiwen},
  booktitle={European Conference on Computer Vision},
  pages={376--393},
  year={2024},
  organization={Springer}
}

@inproceedings{gaussianformer2,
  title={Gaussianformer-2: Probabilistic gaussian superposition for efficient 3d occupancy prediction},
  author={Huang, Yuanhui and Thammatadatrakoon, Amonnut and Zheng, Wenzhao and Zhang, Yunpeng and Du, Dalong and Lu, Jiwen},
  booktitle={Proceedings of the computer vision and pattern recognition conference},
  pages={27477--27486},
  year={2025}
}

@inproceedings{monoscene,
  title={Monoscene: Monocular 3d semantic scene completion},
  author={Cao, Anh-Quan and De Charette, Raoul},
  booktitle={Proceedings of the IEEE/CVF Conference on Computer Vision and Pattern Recognition},
  pages={3991--4001},
  year={2022}
}

@inproceedings{ISO,
  title={Monocular occupancy prediction for scalable indoor scenes},
  author={Yu, Hongxiao and Wang, Yuqi and Chen, Yuntao and Zhang, Zhaoxiang},
  booktitle={European Conference on Computer Vision},
  pages={38--54},
  year={2024},
  organization={Springer}
}

@inproceedings{surroundocc,
  title={Surroundocc: Multi-camera 3d occupancy prediction for autonomous driving},
  author={Wei, Yi and Zhao, Linqing and Zheng, Wenzhao and Zhu, Zheng and Zhou, Jie and Lu, Jiwen},
  booktitle={Proceedings of the IEEE/CVF International Conference on Computer Vision},
  pages={21729--21740},
  year={2023}
}

@inproceedings{vggt,
  title={Vggt: Visual geometry grounded transformer},
  author={Wang, Jianyuan and Chen, Minghao and Karaev, Nikita and Vedaldi, Andrea and Rupprecht, Christian and Novotny, David},
  booktitle={Proceedings of the Computer Vision and Pattern Recognition Conference},
  pages={5294--5306},
  year={2025}
}

@article{pi3,
  title={$\pi^3$: Scalable Permutation-Equivariant Visual Geometry Learning},
  author={Wang, Yifan and Zhou, Jianjun and Zhu, Haoyi and Chang, Wenzheng and Zhou, Yang and Li, Zizun and Chen, Junyi and Pang, Jiangmiao and Shen, Chunhua and He, Tong},
  journal={arXiv preprint arXiv:2507.13347},
  year={2025}
}

@inproceedings{volumetrivln,
  title={Volumetric environment representation for vision-language navigation},
  author={Liu, Rui and Wang, Wenguan and Yang, Yi},
  booktitle={Proceedings of the IEEE/CVF conference on computer vision and pattern recognition},
  pages={16317--16328},
  year={2024}
}

@inproceedings{depthanything,
  title={Depth anything: Unleashing the power of large-scale unlabeled data},
  author={Yang, Lihe and Kang, Bingyi and Huang, Zilong and Xu, Xiaogang and Feng, Jiashi and Zhao, Hengshuang},
  booktitle={Proceedings of the IEEE/CVF conference on computer vision and pattern recognition},
  pages={10371--10381},
  year={2024}
}

@article{depthanythingv2,
  title={Depth anything v2},
  author={Yang, Lihe and Kang, Bingyi and Huang, Zilong and Zhao, Zhen and Xu, Xiaogang and Feng, Jiashi and Zhao, Hengshuang},
  journal={Advances in Neural Information Processing Systems},
  volume={37},
  pages={21875--21911},
  year={2024}
}

@inproceedings{dust3r,
  title={Dust3r: Geometric 3d vision made easy},
  author={Wang, Shuzhe and Leroy, Vincent and Cabon, Yohann and Chidlovskii, Boris and Revaud, Jerome},
  booktitle={Proceedings of the IEEE/CVF Conference on Computer Vision and Pattern Recognition},
  pages={20697--20709},
  year={2024}
}

@inproceedings{mast3r,
  title={Grounding image matching in 3d with mast3r},
  author={Leroy, Vincent and Cabon, Yohann and Revaud, J{\'e}r{\^o}me},
  booktitle={European Conference on Computer Vision},
  pages={71--91},
  year={2024},
  organization={Springer}
}

@inproceedings{voxformer,
  title={Voxformer: Sparse voxel transformer for camera-based 3d semantic scene completion},
  author={Li, Yiming and Yu, Zhiding and Choy, Christopher and Xiao, Chaowei and Alvarez, Jose M and Fidler, Sanja and Feng, Chen and Anandkumar, Anima},
  booktitle={Proceedings of the IEEE/CVF conference on computer vision and pattern recognition},
  pages={9087--9098},
  year={2023}
}

@article{embodiedaisurvey2025,
  title={Embodied intelligence: A synergy of morphology, action, perception and learning},
  author={Liu, Huaping and Guo, Di and Cangelosi, Angelo},
  journal={ACM Computing Surveys},
  volume={57},
  number={7},
  pages={1--36},
  year={2025},
  publisher={ACM New York, NY}
}

@article{bevformer,
  title={Bevformer: learning bird's-eye-view representation from lidar-camera via spatiotemporal transformers},
  author={Li, Zhiqi and Wang, Wenhai and Li, Hongyang and Xie, Enze and Sima, Chonghao and Lu, Tong and Yu, Qiao and Dai, Jifeng},
  journal={IEEE Transactions on Pattern Analysis and Machine Intelligence},
  year={2024},
  publisher={IEEE}
}

@article{occvla,
  title={OccVLA: Vision-Language-Action Model with Implicit 3D Occupancy Supervision},
  author={Liu, Ruixun and Kong, Lingyu and Li, Derun and Zhao, Hang},
  journal={arXiv preprint arXiv:2509.05578},
  year={2025}
}

@article{occllama,
  title={Occllama: An occupancy-language-action generative world model for autonomous driving},
  author={Wei, Julong and Yuan, Shanshuai and Li, Pengfei and Hu, Qingda and Gan, Zhongxue and Ding, Wenchao},
  journal={arXiv preprint arXiv:2409.03272},
  year={2024}
}

@inproceedings{occupancypoints,
  title={Occupancy as set of points},
  author={Shi, Yiang and Cheng, Tianheng and Zhang, Qian and Liu, Wenyu and Wang, Xinggang},
  booktitle={European Conference on Computer Vision},
  pages={72--87},
  year={2024},
  organization={Springer}
}

@article{opus,
  title={Opus: occupancy prediction using a sparse set},
  author={Wang, Jiabao and Liu, Zhaojiang and Meng, Qiang and Yan, Liujiang and Wang, Ke and Yang, Jie and Liu, Wei and Hou, Qibin and Cheng, Ming-Ming},
  journal={Advances in Neural Information Processing Systems},
  volume={37},
  pages={119861--119885},
  year={2024}
}

@inproceedings{lss,
  title={Lift, splat, shoot: Encoding images from arbitrary camera rigs by implicitly unprojecting to 3d},
  author={Philion, Jonah and Fidler, Sanja},
  booktitle={European conference on computer vision},
  pages={194--210},
  year={2020},
  organization={Springer}
}

@inproceedings{fast3r,
  title={Fast3r: Towards 3d reconstruction of 1000+ images in one forward pass},
  author={Yang, Jianing and Sax, Alexander and Liang, Kevin J and Henaff, Mikael and Tang, Hao and Cao, Ang and Chai, Joyce and Meier, Franziska and Feiszli, Matt},
  booktitle={Proceedings of the Computer Vision and Pattern Recognition Conference},
  pages={21924--21935},
  year={2025}
}

@article{point3r,
  title={Point3r: Streaming 3d reconstruction with explicit spatial pointer memory},
  author={Wu, Yuqi and Zheng, Wenzhao and Zhou, Jie and Lu, Jiwen},
  journal={Advances in Neural Information Processing Systems},
  volume={38},
  pages={69675--69699},
  year={2026}
}

@article{adamw,
  title={Decoupled weight decay regularization},
  author={Loshchilov, Ilya and Hutter, Frank},
  journal={arXiv preprint arXiv:1711.05101},
  year={2017}
}

@inproceedings{Triformer,
  title={Tri-perspective view for vision-based 3d semantic occupancy prediction},
  author={Huang, Yuanhui and Zheng, Wenzhao and Zhang, Yunpeng and Zhou, Jie and Lu, Jiwen},
  booktitle={Proceedings of the IEEE/CVF conference on computer vision and pattern recognition},
  pages={9223--9232},
  year={2023}
}

@inproceedings{imvoxelnet,
  title={Imvoxelnet: Image to voxels projection for monocular and multi-view general-purpose 3d object detection},
  author={Rukhovich, Danila and Vorontsova, Anna and Konushin, Anton},
  booktitle={Proceedings of the IEEE/CVF winter conference on applications of computer vision},
  pages={2397--2406},
  year={2022}
}

@inproceedings{indoordepth,
  title={Toward practical monocular indoor depth estimation},
  author={Wu, Cho-Ying and Wang, Jialiang and Hall, Michael and Neumann, Ulrich and Su, Shuochen},
  booktitle={Proceedings of the IEEE/CVF conference on computer vision and pattern recognition},
  pages={3814--3824},
  year={2022}
}

@inproceedings{unimode,
  title={Unimode: Unified monocular 3d object detection},
  author={Li, Zhuoling and Xu, Xiaogang and Lim, SerNam and Zhao, Hengshuang},
  booktitle={Proceedings of the IEEE/CVF Conference on Computer Vision and Pattern Recognition},
  pages={16561--16570},
  year={2024}
}

@inproceedings{occfor3ddet,
  title={Learning occupancy for monocular 3d object detection},
  author={Peng, Liang and Xu, Junkai and Cheng, Haoran and Yang, Zheng and Wu, Xiaopei and Qian, Wei and Wang, Wenxiao and Wu, Boxi and Cai, Deng},
  booktitle={Proceedings of the IEEE/CVF Conference on Computer Vision and Pattern Recognition},
  pages={10281--10292},
  year={2024}
}

@inproceedings{nuscenes,
  title={nuscenes: A multimodal dataset for autonomous driving},
  author={Caesar, Holger and Bankiti, Varun and Lang, Alex H and Vora, Sourabh and Liong, Venice Erin and Xu, Qiang and Krishnan, Anush and Pan, Yu and Baldan, Giancarlo and Beijbom, Oscar},
  booktitle={Proceedings of the IEEE/CVF conference on computer vision and pattern recognition},
  pages={11621--11631},
  year={2020}
}

@inproceedings{ndcscene,
  title={Ndc-scene: Boost monocular 3d semantic scene completion in normalized device coordinates space},
  author={Yao, Jiawei and Li, Chuming and Sun, Keqiang and Cai, Yingjie and Li, Hao and Ouyang, Wanli and Li, Hongsheng},
  booktitle={2023 IEEE/CVF International Conference on Computer Vision (ICCV)},
  pages={9421--9431},
  year={2023},
  organization={IEEE Computer Society}
}

@inproceedings{sscnet,
  title={Semantic scene completion from a single depth image},
  author={Song, Shuran and Yu, Fisher and Zeng, Andy and Chang, Angel X and Savva, Manolis and Funkhouser, Thomas},
  booktitle={Proceedings of the IEEE conference on computer vision and pattern recognition},
  pages={1746--1754},
  year={2017}
}

@inproceedings{sparseocc,
  title={Sparseocc: Rethinking sparse latent representation for vision-based semantic occupancy prediction},
  author={Tang, Pin and Wang, Zhongdao and Wang, Guoqing and Zheng, Jilai and Ren, Xiangxuan and Feng, Bailan and Ma, Chao},
  booktitle={Proceedings of the IEEE/CVF Conference on Computer Vision and Pattern Recognition},
  pages={15035--15044},
  year={2024}
}

@article{fbocc,
  title={{FB-OCC}: {3D} Occupancy Prediction based on Forward-Backward View Transformation},
  author={Li, Zhiqi and Yu, Zhiding and Austin, David and Fang, Mingsheng and Lan, Shiyi and Kautz, Jan and Alvarez, Jose M},
  journal={arXiv:2307.01492},
  year={2023}
}

@article{octreeocc,
  title={Octreeocc: Efficient and multi-granularity occupancy prediction using octree queries},
  author={Lu, Yuhang and Zhu, Xinge and Wang, Tai and Ma, Yuexin},
  journal={Advances in Neural Information Processing Systems},
  volume={37},
  pages={79618--79641},
  year={2024}
}

@article{streamvggt,
  title={Streaming 4d visual geometry transformer},
  author={Zhuo, Dong and Zheng, Wenzhao and Guo, Jiahe and Wu, Yuqi and Zhou, Jie and Lu, Jiwen},
  journal={arXiv preprint arXiv:2507.11539},
  year={2025}
}

@article{dens3r,
  title={Dens3R: A Foundation Model for 3D Geometry Prediction},
  author={Fang, Xianze and Gao, Jingnan and Wang, Zhe and Chen, Zhuo and Ren, Xingyu and Lyu, Jiangjing and Ren, Qiaomu and Yang, Zhonglei and Yang, Xiaokang and Yan, Yichao and others},
  journal={arXiv preprint arXiv:2507.16290},
  year={2025}
}

@InProceedings{cut3r,
    author    = {Wang, Qianqian and Zhang, Yifei and Holynski, Aleksander and Efros, Alexei A. and Kanazawa, Angjoo},
    title     = {Continuous 3D Perception Model with Persistent State},
    booktitle = {Proceedings of the IEEE/CVF Conference on Computer Vision and Pattern Recognition (CVPR)},
    month     = {June},
    year      = {2025},
    pages     = {10510-10522}
}

@InProceedings{must3r,
    author    = {Cabon, Yohann and Stoffl, Lucas and Antsfeld, Leonid and Csurka, Gabriela and Chidlovskii, Boris and Revaud, Jerome and Leroy, Vincent},
    title     = {MUSt3R: Multi-view Network for Stereo 3D Reconstruction},
    booktitle = {Proceedings of the IEEE/CVF Conference on Computer Vision and Pattern Recognition (CVPR)},
    month     = {June},
    year      = {2025},
    pages     = {1050-1060}
}

@inproceedings{spann3r,
  title={3d reconstruction with spatial memory},
  author={Wang, Hengyi and Agapito, Lourdes},
  booktitle={2025 International Conference on 3D Vision (3DV)},
  pages={78--89},
  year={2025},
  organization={IEEE}
}

@inproceedings{clip,
  title={Learning transferable visual models from natural language supervision},
  author={Radford, Alec and Kim, Jong Wook and Hallacy, Chris and Ramesh, Aditya and Goh, Gabriel and Agarwal, Sandhini and Sastry, Girish and Askell, Amanda and Mishkin, Pamela and Clark, Jack and others},
  booktitle={International conference on machine learning},
  pages={8748--8763},
  year={2021},
  organization={PmLR}
}

@InProceedings{OccFormer,
    author    = {Zhang, Yunpeng and Zhu, Zheng and Du, Dalong},
    title     = {OccFormer: Dual-path Transformer for Vision-based 3D Semantic Occupancy Prediction},
    booktitle = {Proceedings of the IEEE/CVF International Conference on Computer Vision (ICCV)},
    month     = {October},
    year      = {2023},
    pages     = {9433-9443}
}

@article{roboocc,
  title={Roboocc: Enhancing the geometric and semantic scene understanding for robots},
  author={Zhang, Zhang and Zhang, Qiang and Cui, Wei and Shi, Shuai and Guo, Yijie and Han, Gang and Zhao, Wen and Ren, Hengle and Xu, Renjing and Tang, Jian},
  journal={arXiv preprint arXiv:2504.14604},
  year={2025}
}

@inproceedings{atlas,
  title={Atlas: End-to-end 3d scene reconstruction from posed images},
  author={Murez, Zak and As, Tarrence van and Bartolozzi, James and Sinha, Ayan and Badrinarayanan, Vijay and Rabinovich, Andrew},
  booktitle={ECCV},
  pages={414--431},
  year={2020}
}

@inproceedings{gpocc,
  title={Generalizing Visual Geometry Priors to Sparse Gaussian Occupancy Prediction},
  author={Zhou, Changqing and Luo, Yueru and Chen, Changhao},
  booktitle={Proceedings of the IEEE/CVF Conference on Computer Vision and Pattern Recognition},
  pages={28578--28587},
  year={2026}
}

@inproceedings{legoocc,
  title={Monocular open vocabulary occupancy prediction for indoor scenes},
  author={Zhou, Changqing and Luo, Yueru and Zhang, Han and Jiang, Zeyu and Chen, Changhao},
  booktitle={Proceedings of the IEEE/CVF Conference on Computer Vision and Pattern Recognition},
  pages={21627--21637},
  year={2026}
}

@article{zuo2025quadricformer,
  title={Quadricformer: Scene as superquadrics for 3d semantic occupancy prediction},
  author={Zuo, Sicheng and Zheng, Wenzhao and Han, Xiaoyong and Yang, Longchao and Lu, Jiwen and others},
  journal={Advances in Neural Information Processing Systems},
  volume={38},
  pages={47779--47801},
  year={2026}
}

@article{plucker1865xvii,
  title={Xvii. on a new geometry of space},
  author={Plucker, Julius},
  journal={Philosophical Transactions of the Royal Society of London},
  number={155},
  pages={725--791},
  year={1865},
  publisher={The Royal Society London}
}

@inproceedings{jiang2025rayzer,
  title={Rayzer: A self-supervised large view synthesis model},
  author={Jiang, Hanwen and Tan, Hao and Wang, Peng and Jin, Haian and Zhao, Yue and Bi, Sai and Zhang, Kai and Luan, Fujun and Sunkavalli, Kalyan and Huang, Qixing and others},
  booktitle={Proceedings of the IEEE/CVF International Conference on Computer Vision},
  pages={4918--4929},
  year={2025}
}

@article{shazeer2019fast,
  title={Fast transformer decoding: One write-head is all you need},
  author={Shazeer, Noam},
  journal={arXiv preprint arXiv:1911.02150},
  year={2019}
}

@article{jiang2026freeocc,
  title={FreeOcc: Training-Free Embodied Open-Vocabulary Occupancy Prediction},
  author={Jiang, Zeyu and Zhou, Changqing and Zuo, Xingxing and Chen, Changhao},
  journal={arXiv preprint arXiv:2604.28115},
  year={2026}
}

@article{li2025occscene,
  title={OccScene: Semantic occupancy-based cross-task mutual learning for 3D scene generation},
  author={Li, Bohan and Jin, Xin and Wang, Jianan and Shi, Yukai and Sun, Yasheng and Wang, Xiaofeng and Ma, Zhuang and Xie, Baao and Ma, Chao and Yang, Xiaokang and others},
  journal={IEEE Transactions on Pattern Analysis and Machine Intelligence},
  year={2025},
  publisher={IEEE}
}

@article{yan2025spot,
  title={SPOT: Scalable 3D pre-training via occupancy prediction for learning transferable 3D representations},
  author={Yan, Xiangchao and Chen, Runjian and Zhang, Bo and Ye, Hancheng and Xia, Renqiu and Yuan, Jiakang and Zhou, Hongbin and Cai, Xinyu and Shi, Botian and Shao, Wenqi and others},
  journal={IEEE Transactions on Pattern Analysis and Machine Intelligence},
  year={2025},
  publisher={IEEE}
}

@article{rist2021semantic,
  title={Semantic scene completion using local deep implicit functions on lidar data},
  author={Rist, Christoph B and Emmerichs, David and Enzweiler, Markus and Gavrila, Dariu M},
  journal={IEEE transactions on pattern analysis and machine intelligence},
  volume={44},
  number={10},
  pages={7205--7218},
  year={2021},
  publisher={IEEE}
}

@article{li2021anisotropic,
  title={Anisotropic convolutional neural networks for RGB-D based semantic scene completion},
  author={Li, Jie and Wang, Peng and Han, Kai and Liu, Yu},
  journal={IEEE Transactions on Pattern Analysis and Machine Intelligence},
  volume={44},
  number={11},
  pages={8125--8138},
  year={2021},
  publisher={IEEE}
}

@article{huang2024ssr,
  title={Ssr-2d: semantic 3d scene reconstruction from 2d images},
  author={Huang, Junwen and Artemov, Alexey and Chen, Yujin and Zhi, Shuaifeng and Xu, Kai and Nie{\ss}ner, Matthias},
  journal={IEEE Transactions on Pattern Analysis and Machine Intelligence},
  volume={46},
  number={12},
  pages={8486--8501},
  year={2024},
  publisher={IEEE}
}

@article{zheng2026omnihd,
  title={OmniHD-Scenes: A next-generation multimodal dataset for autonomous driving},
  author={Zheng, Lianqing and Yang, Long and Lin, Qunshu and Ai, Wenjin and Liu, Minghao and Lu, Shouyi and Liu, Jianan and Ren, Hongze and Mo, Jingyue and Bai, Xiaokai and others},
  journal={IEEE Transactions on Pattern Analysis and Machine Intelligence},
  year={2026},
  publisher={IEEE}
}

@article{liu2025hybrid,
  title={Hybrid-prediction integrated planning for autonomous driving},
  author={Liu, Haochen and Huang, Zhiyu and Huang, Wenhui and Yang, Haohan and Mo, Xiaoyu and Lv, Chen},
  journal={IEEE Transactions on Pattern Analysis and Machine Intelligence},
  volume={47},
  number={4},
  pages={2597--2614},
  year={2025},
  publisher={IEEE}
}

@article{ranftl2020towards,
  title={Towards robust monocular depth estimation: Mixing datasets for zero-shot cross-dataset transfer},
  author={Ranftl, Ren{\'e} and Lasinger, Katrin and Hafner, David and Schindler, Konrad and Koltun, Vladlen},
  journal={IEEE transactions on pattern analysis and machine intelligence},
  volume={44},
  number={3},
  pages={1623--1637},
  year={2020},
  publisher={IEEE}
}

@article{li2023delving,
  title={Delving into the devils of bird’s-eye-view perception: A review, evaluation and recipe},
  author={Li, Hongyang and Sima, Chonghao and Dai, Jifeng and Wang, Wenhai and Lu, Lewei and Wang, Huijie and Zeng, Jia and Li, Zhiqi and Yang, Jiazhi and Deng, Hanming and others},
  journal={IEEE Transactions on Pattern Analysis and Machine Intelligence},
  volume={46},
  number={4},
  pages={2151--2170},
  year={2023},
  publisher={IEEE}
}

@article{shi2023raymvsnet++,
  title={RayMVSNet++: learning ray-based 1D implicit fields for accurate multi-view stereo},
  author={Shi, Yifei and Xi, Junhua and Hu, Dewen and Cai, Zhiping and Xu, Kai},
  journal={IEEE Transactions on Pattern Analysis and Machine Intelligence},
  volume={45},
  number={11},
  pages={13666--13682},
  year={2023},
  publisher={IEEE}
}

@article{ma2024vision,
  title={Vision-centric bev perception: A survey},
  author={Ma, Yuexin and Wang, Tai and Bai, Xuyang and Yang, Huitong and Hou, Yuenan and Wang, Yaming and Qiao, Yu and Yang, Ruigang and Zhu, Xinge},
  journal={IEEE Transactions on Pattern Analysis and Machine Intelligence},
  volume={46},
  number={12},
  pages={10978--10997},
  year={2024},
  publisher={IEEE}
}

@article{yang2025bevheight++,
  title={BEVHeight++: Toward robust visual centric 3D object detection},
  author={Yang, Lei and Tang, Tao and Li, Jun and Yuan, Kun and Wu, Kai and Chen, Peng and Wang, Li and Huang, Yi and Li, Lei and Zhang, Xinyu and others},
  journal={IEEE Transactions on Pattern Analysis and Machine Intelligence},
  year={2025},
  publisher={IEEE}
}

@article{chen2020visibility,
  title={Visibility-aware point-based multi-view stereo network},
  author={Chen, Rui and Han, Songfang and Xu, Jing and Su, Hao},
  journal={IEEE transactions on pattern analysis and machine intelligence},
  volume={43},
  number={10},
  pages={3695--3708},
  year={2020},
  publisher={IEEE}
}

@article{lee2025visibility,
  title={Visibility-Aware Multi-View Stereo by Surface Normal Weighting for Occlusion Robustness},
  author={Lee, Hyucksang and Lee, Seongmin and Lee, Sanghoon},
  journal={IEEE transactions on pattern analysis and machine intelligence},
  year={2025},
  publisher={IEEE}
}
}

\begin{IEEEbiography}[{\includegraphics[width=1in,height=1.25in,clip,keepaspectratio]{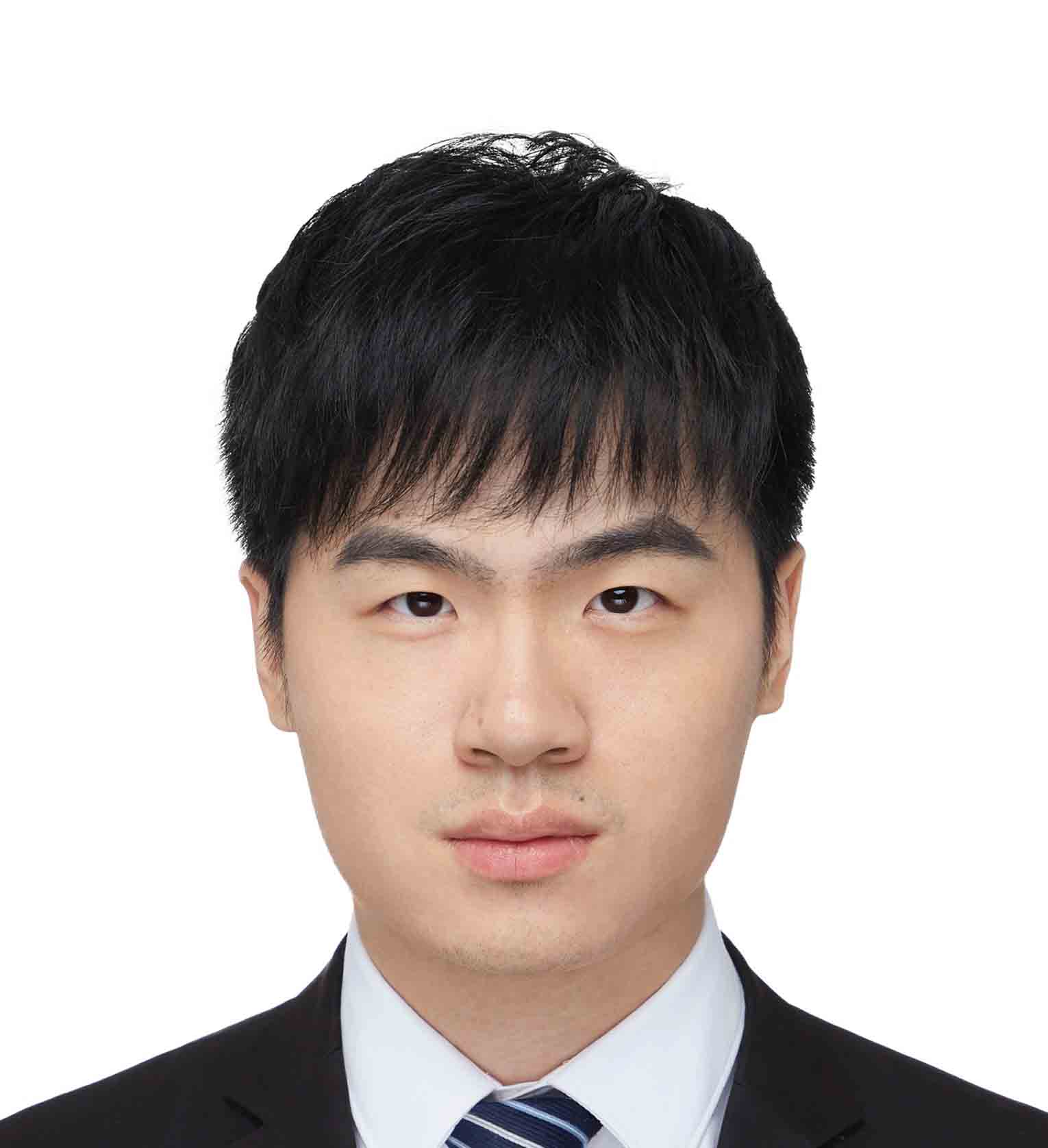}}]
{Changqing Zhou} is currently pursuing the Ph.D. degree at The Hong Kong University of Science and Technology (Guangzhou). His research interests include 3D computer vision, embodied scene understanding, occupancy prediction, and open-vocabulary 3D perception. His recent work focuses on 3D scene understanding and world modeling for visual navigation.
\end{IEEEbiography}

\begin{IEEEbiography}[{\includegraphics[width=1in,height=1.25in,clip,keepaspectratio]{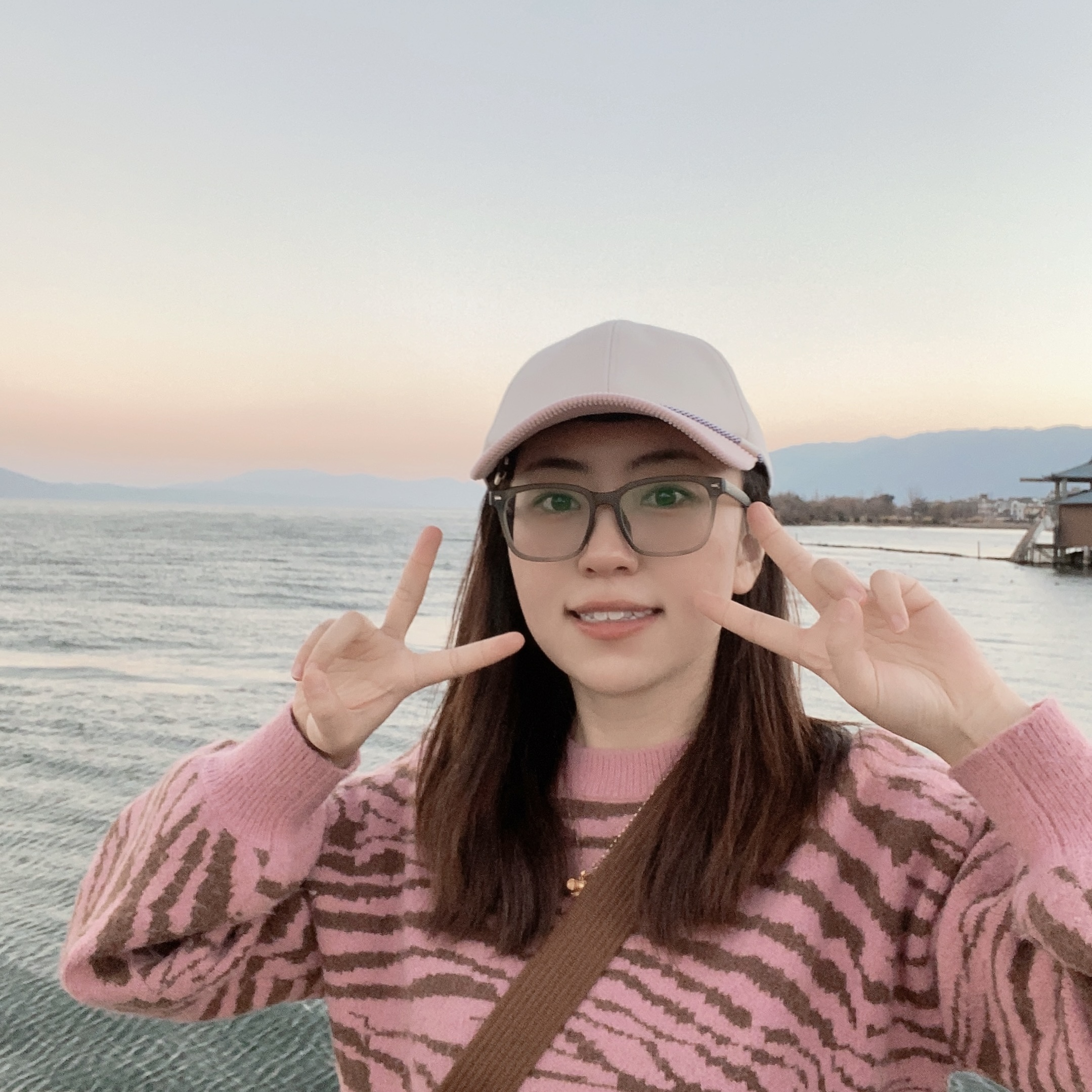}}]
{Yueru Luo} is currently pursuing the Ph.D. degree at The Chinese University of Hong Kong, Shenzhen. Her research interests include 3D computer vision, autonomous-driving perception, 3D lane detection, scene topology reasoning and scene understanding in autonomous-driving environments. Her recent work focuses on large-scale 4D geometry foundation models and vision-language models.
\end{IEEEbiography}

\begin{IEEEbiography}[{\includegraphics[width=1in,height=1.25in,clip,keepaspectratio]{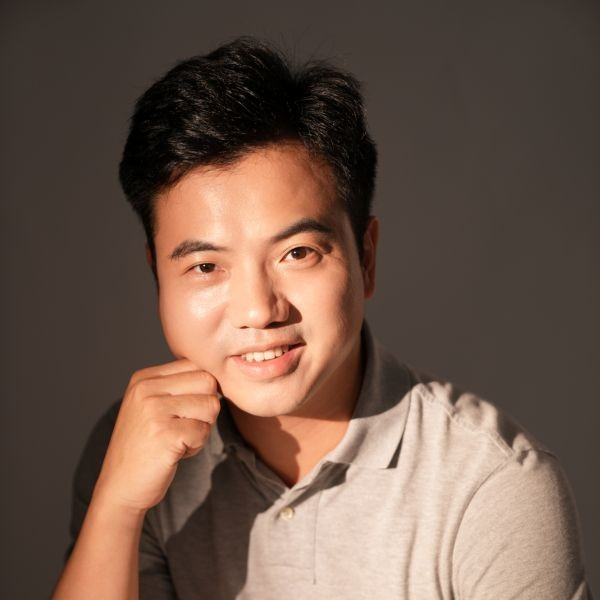}}]
{Yulan Guo} is a full Professor with the School of Electronics and Communication Engineering, Sun Yat-sen University. His research interests lie in spatial intelligence, embodied intelligence, and 3D vision. He has authored over 200 articles at highly referred journals and conferences. He served as a Senior Area Editor for IEEE Transactions on Image Processing, and an Associate Editor for the Visual Computer, and Computers \& Graphics. He also served as an area chair for CVPR, ICCV, ECCV, NeurIPS, ICML, and ACM Multimedia. He organized over 10 workshops, challenges, and tutorials in prestigious conferences such as CVPR, ICCV, ECCV, and 3DV. He is a Senior Member of IEEE and ACM.
\end{IEEEbiography}

\begin{IEEEbiography}[{\includegraphics[width=1in,height=1.25in,clip,keepaspectratio]{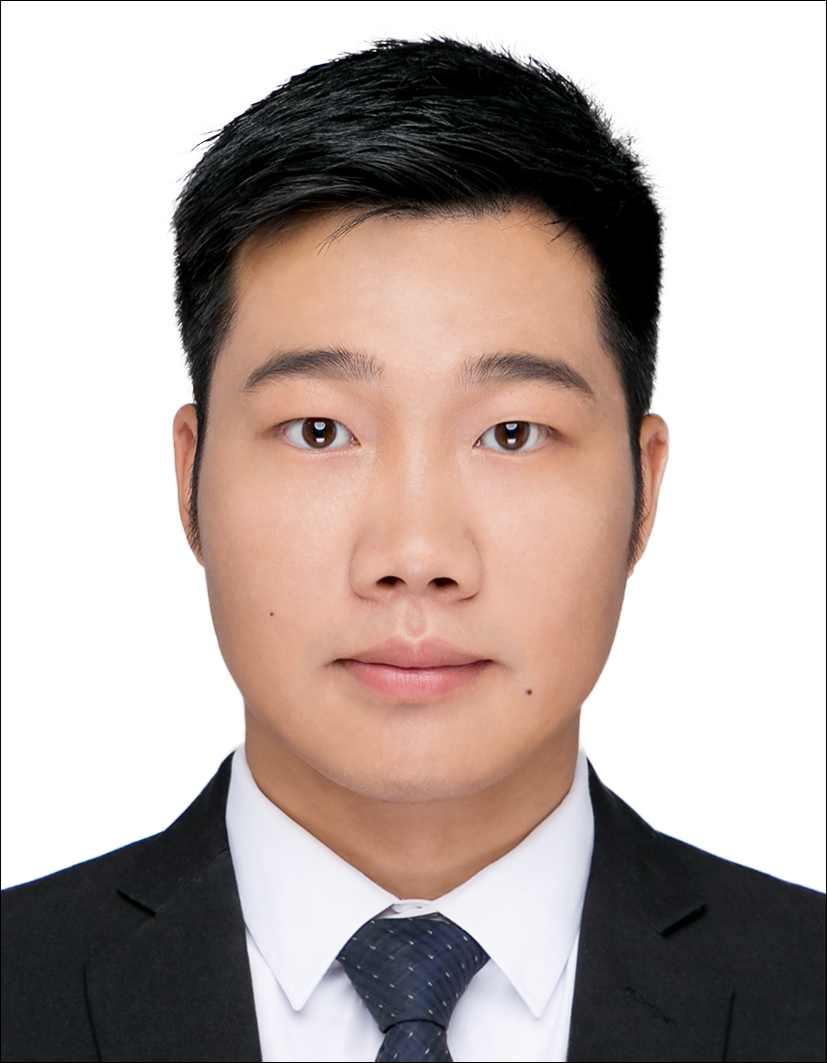}}]
{Bing Wang} is an Assistant Professor in Robotics and Autonomous Systems at the Faculty of Engineering, The Hong Kong Polytechnic University. He obtained his DPhil degree in 2022 from the Department of Computer Science at the University of Oxford. His research is at the forefront of spatial intelligence, a dynamic field focused on advancing human-level 3D spatial perception and world understanding for mobile robotics. The primary objective is to enhance the reliability, intelligence, and security of intelligent machines in real-world environments. He is serving/served as an Associate Editor of IEEE TIP, Science Bulletin and a Technical Papers Committee of ACM SIGGRAPH Asia.
\end{IEEEbiography}

\begin{IEEEbiography}[{\includegraphics[width=1in,height=1.25in,clip,keepaspectratio]{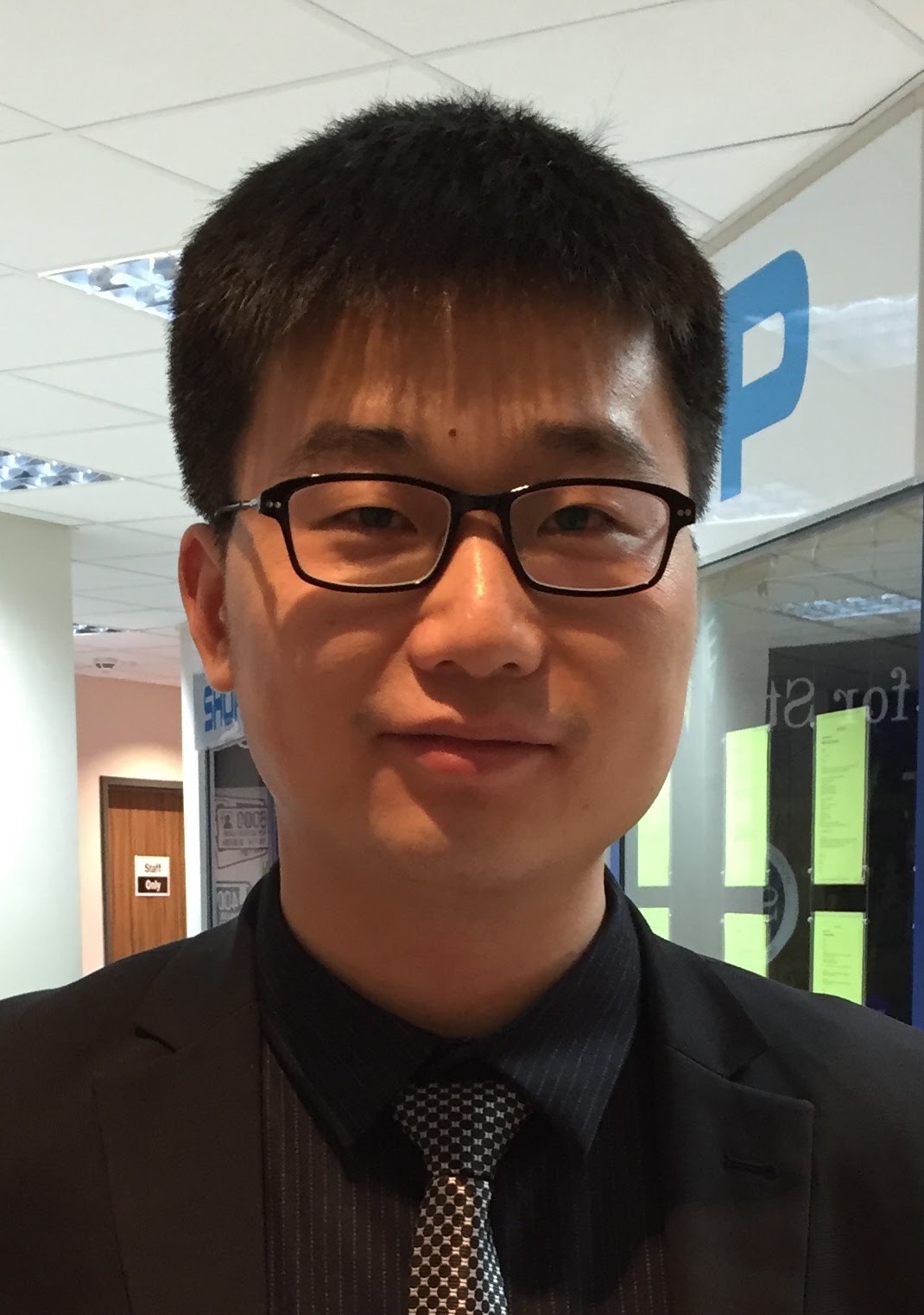}}]
{Jie Qin} is currently a Professor at Nanjing University of Aeronautics and Astronautics, China. He received the B.E. and Ph.D. degrees from Beihang University, China, in 2011 and 2017, respectively. His current research interests include computer vision and machine learning. He has published over 100 papers in top-tier journals/conferences, including IEEE TPAMI, IJCV, CVPR, ICCV, ECCV, AAAI, IJCAI, ICML and NeurIPS. He is serving/served as Associate Editors of IEEE TIP and Neural Networks, a Guest Editor of IJCV, a Program Chair of an ECCV Workshop, Senior PC members of AAAI and IJCAI, and Area Chairs of ICLR, ICML, NeurIPS, ACM MM, ECAI and ICME.
\end{IEEEbiography}

\begin{IEEEbiography}[{\includegraphics[width=1in,height=1.25in,clip,keepaspectratio]{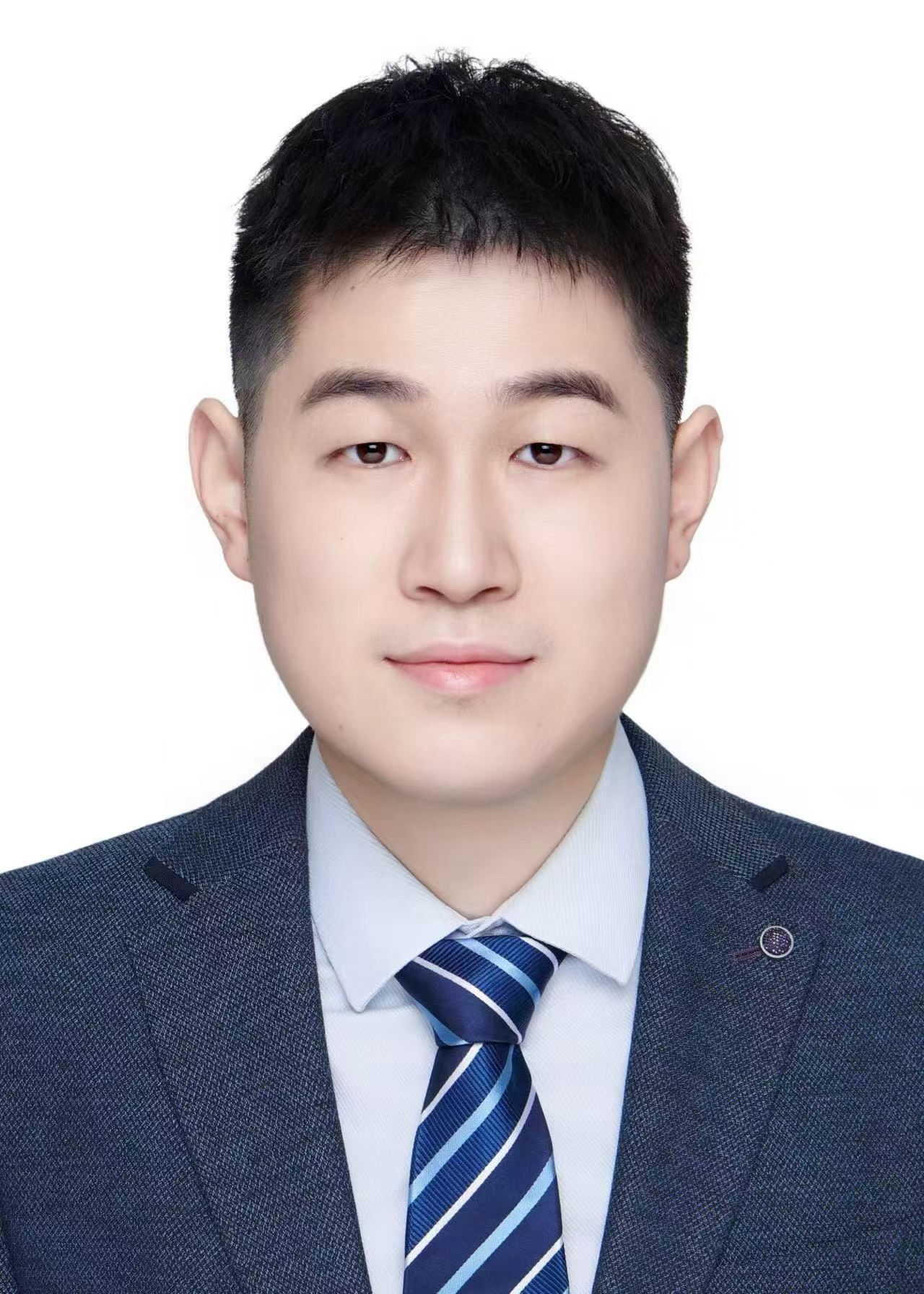}}]
{Changhao Chen} obtained his Ph.D. degree at University of Oxford (UK), M.Eng. degree at the National University of Defense Technology (China), and B.Eng. degree at Tongji University (China). Now he is an Assistant Professor at the Thrust of Intelligent Transportation and Thrust of Artificial Intelligence, the Hong Kong University of Science and Technology (Guangzhou). His research interest lies in robotics, embodied AI and cyber-physical systems.
\end{IEEEbiography}

\end{document}